\pdfoutput=1

\documentclass[11pt]{article}

\usepackage[final]{acl}

\usepackage{times}
\usepackage{latexsym}
\usepackage{booktabs}
\usepackage{tikz}
\usepackage{amsmath}
\usepackage{threeparttable}
\usepackage{filecontents}

\usepackage{amssymb}
\usepackage{amsfonts}
\usepackage{amsthm}
\usepackage{algorithm}
\usepackage{algpseudocode}
\usepackage{textcomp}
\usepackage{xcolor}
\usepackage{soul}
\usepackage{enumitem}
\usepackage{multirow}
\usepackage{graphicx}
\usepackage{subcaption}
\usepackage{longtable}
\usepackage{tabulary}

\usepackage{CJKutf8}
\newcommand{\zh}[1]{\begin{CJK}{UTF8}{gbsn}#1\end{CJK}}
\newcommand{\jp}[1]{\begin{CJK}{UTF8}{min}#1\end{CJK}}

\usepackage[T1]{fontenc}

\usepackage[utf8]{inputenc}

\usepackage{microtype}

\usepackage{inconsolata}

\usepackage{graphicx}

\usepackage{tcolorbox}
\tcbuselibrary{breakable}
\usepackage{colortbl}
\usepackage{adjustbox}

\title{CultureSynth: A Hierarchical Taxonomy-Guided and Retrieval-Augmented Framework for Cultural Question-Answer Synthesis}

\author{Xinyu Zhang\thanks{Corresponding author},
	Pei Zhang, 
        Shuang Luo,
        Jialong Tang,
        Yu Wan,
        Baosong Yang,
        Fei Huang
 \\ 
    Tongyi Lab, Alibaba Group Inc\\ 
	\texttt{\{zxy440266, xiaoyi.zp, shuangluo.ls, tangjialong.tjl\}@alibaba-inc.com} \\
        \texttt{\{wanyu.wy, yangbaosong.ybs, f.huang\}@alibaba-inc.com}
 }
 
\begin{document}
\maketitle

\begin{abstract}


Cultural competence, defined as the ability to understand and adapt to multicultural contexts, is increasingly vital for large language models (LLMs) in global environments. While several cultural benchmarks exist to assess LLMs' cultural competence, current evaluations suffer from fragmented taxonomies, domain specificity, and heavy reliance on manual data annotation. To address these limitations, we introduce CultureSynth, a novel framework comprising (1) a comprehensive hierarchical multilingual cultural taxonomy covering 12 primary and 130 secondary topics, and (2) a Retrieval-Augmented Generation (RAG)-based methodology leveraging factual knowledge to synthesize culturally relevant question-answer pairs. The CultureSynth-7 synthetic benchmark contains 19,360 entries and 4,149 manually verified entries across 7 languages. Evaluation of 14 prevalent LLMs of different sizes reveals clear performance stratification led by ChatGPT-4o-Latest and Qwen2.5-72B-Instruct. The results demonstrate that a 3B-parameter threshold is necessary for achieving basic cultural competence, models display varying architectural biases in knowledge processing, and significant geographic disparities exist across models. We believe that CultureSynth offers a scalable framework for developing culturally aware AI systems while reducing reliance on manual annotation\footnote{Benchmark is available at \url{https://github.com/Eyr3/CultureSynth}.}.





\end{abstract}

\section{Introduction}

Cultural competence is a critical component of complex human emotional intelligence~\cite{goleman1998working}, encompassing key abilities such as empathy, cognition, and adaptability in understanding and navigating cultural differences~\cite{earley2003cultural, earley2004cultural}. For individuals, cultural competence is not only demonstrated through awareness and understanding of cultural knowledge but also through the ability to adapt behavior and communication styles within multicultural contexts~\cite{bennett2004becoming, hong2023multicultural}. 
As the utilization of large language models (LLMs) in interactions, communications~\cite{achiam2023gpt, yang2024qwen2}, and workflows~\cite{gao2024collabcoder,zhang2024aflow} within global environments continue to grow, it is imperative to enhance the cultural competence of these models to optimize performance~\cite{kasneci2023chatgpt}. This enhancement enables LLMs to more effectively interact with individuals from diverse cultural backgrounds worldwide~\cite{havaldar2023multilingual, liu2024multilingual}.

\begin{figure*}[t]
  \centering
  \includegraphics[width=\linewidth]{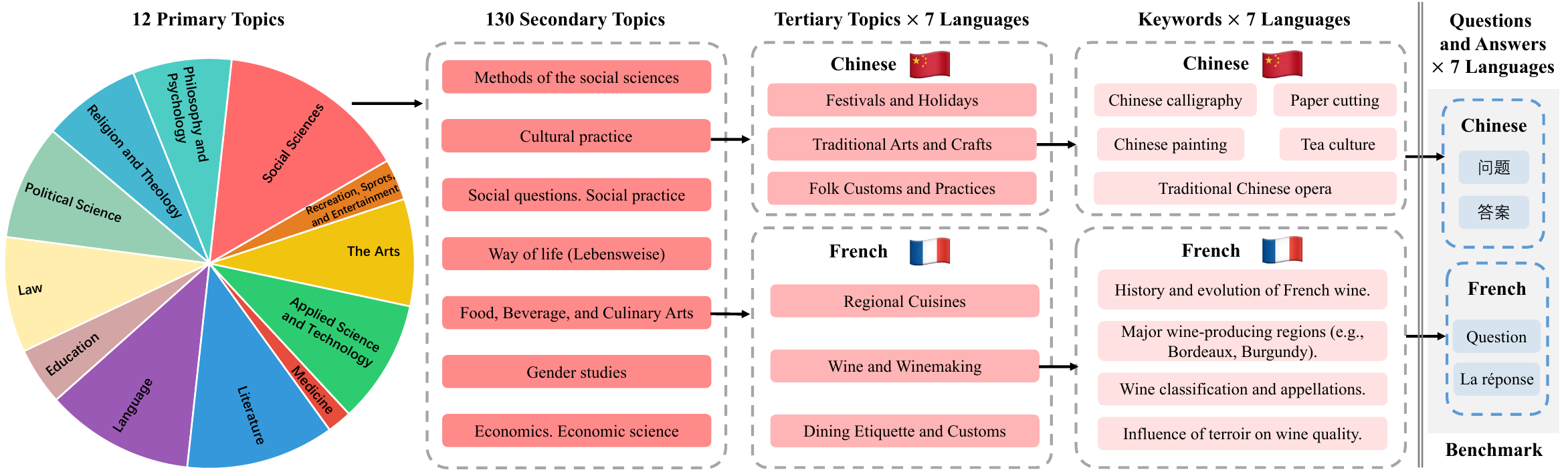}
  \vspace{-4mm}
  \caption{CultureSynth consists of a hierarchically structured multilingual cultural taxonomy (left of double line) and a RAG-based question-answer synthesis methodology (right),  which together generate the benchmark.}
  \label{fig:intro}
  \vspace{-4mm}
\end{figure*}


Several benchmarks have been developed to evaluate the cultural competence of LLMs, primarily using multiple-choice and question-answer (QA) generation formats~\cite{pawar2024survey,minaee2024large}. Multiple-choice benchmarks such as BLEnD~\cite{myung2024blend}, CommonsenseQA~\cite{putri2024can}, and MMLU~\cite{hendrycksmeasuring} gather culturally specific questions and manually annotate the correct and incorrect options. Generation benchmarks like CaLMQA~\cite{arora2024calmqa}, NativQA~\cite{hasan2024nativqa}, and CultureBank~\cite{shi2024culturebank} compile naturally occurring questions from Wikipedia and community web forums~\cite{fan2019eli5}, in addition to human-written questions annotated by native speakers. Other benchmarks, such as PRISM~\cite{kirk2024prism}, cluster and filter culturally specific queries from real-world inquiries sourced via an English-speaking participant crowdwork platform. Meanwhile, CulturePark~\cite{li2024culturepark} employs a different approach, creating a cultural benchmark through dialogues between LLM-based multi-agent communication.









However, existing cultural benchmarks encounter two primary challenges. First, cultural competence spans a wide range of domains (e.g., religion, social customs, law), yet most current benchmarks address only specific discrete cultural topics and lack a systematic taxonomy~\cite{myung2024blend,chiu2024culturalteaming}. This gap hinders comprehensive and systematic analyses of LLMs' cultural capabilities across varied domains. Second, these benchmarks mainly rely on existing forums and manual annotation~\cite{arora2024calmqa,shi2024culturebank}. While such benchmark construction methods ensure quality, they are resource-intensive, requiring substantial human labor and material costs, and cannot guarantee comprehensiveness. Moreover, as the consumption of training data accelerates, LLMs increasingly face data shortages~\cite{villalobos2024position}. Consequently, employing automated methods to synthesize high-quality data becomes essential~\cite{long2024llms}.

To address the aforementioned challenges, we introduce CultureSynth, as illustrated in Figure~\ref{fig:intro}, a novel cultural framework featuring an automated data synthesis methodology. We propose the first hierarchically structured multilingual cultural taxonomy to tackle the issue of dispersed and unsystematic cultural topics. This framework integrates library classifications from five countries and regions to create a universal cultural taxonomy comprising 12 primary topics and 130 secondary topics (see Table~\ref{tab:topics_all}), encompassing diverse global cultures. Additionally, we employ an expert role-playing LLM to extend over a thousand deep, country-specific cultural topics for each language. To further construct a reliable cultural benchmark while minimizing manual annotation, we propose a Retrieval-Augmented Generation (RAG)-based methodology for question-answer pair synthesis. Leveraging carefully vetted factual knowledge that ensures data authenticity, we employ expert role-playing and instruction-following prompts to generate safe, clear, and culturally relevant questions with high-quality answers. 



The CultureSynth-7 benchmark encompasses 7 languages (Arabic [ar], Spanish [es], French [fr], Japanese [ja], Korean [ko], Portuguese [pt], and Chinese [zh]), totaling 19,360 QA pairs with thousands of cultural keywords per language.
Native speaker annotation of a representative subset validated the benchmark quality, achieving 95.8\% question clarity, 83.5\% cultural relevance, and 98.8\% answer quality, with no safety concerns identified.



Using CultureSynth, we conduct an extensive evaluation across 14 widely used language models, including GPT-4o~\cite{achiam2023gpt}, Claude-3.5~\cite{anthropic2024claude35}, Qwen2.5~\cite{yang2024qwen2}, Llama 3~\cite{dubey2024llama}, and the Mistral~\cite{jiang2024mixtral} series. 
The evaluation reveals a clear performance hierarchy among LLMs, i.e., ChatGPT-4o-Latest $>$ Qwen2.5-72B $>$ Claude-3.5-Sonnet $>$ others. 
We identify a 3B-parameter threshold for basic cultural competence, below which models default to native-language functionality. Model architecture plays a crucial role: mixture-of-experts architectures excel in cultural knowledge retrieval, while dense transformers perform better in long-context scenarios. Our evaluation also uncovers distinct geographic and domain-specific biases, particularly GPT-4o's performance gaps in East Asian contexts and Claude-3.5's limitations in Arabic and Korean language processing. 

\section{Related Work}
\subsection{Cultural Competence Benchmark}
\vspace{-1mm}

\begin{figure*}[!t]
  \centering
  \includegraphics[width=\linewidth]{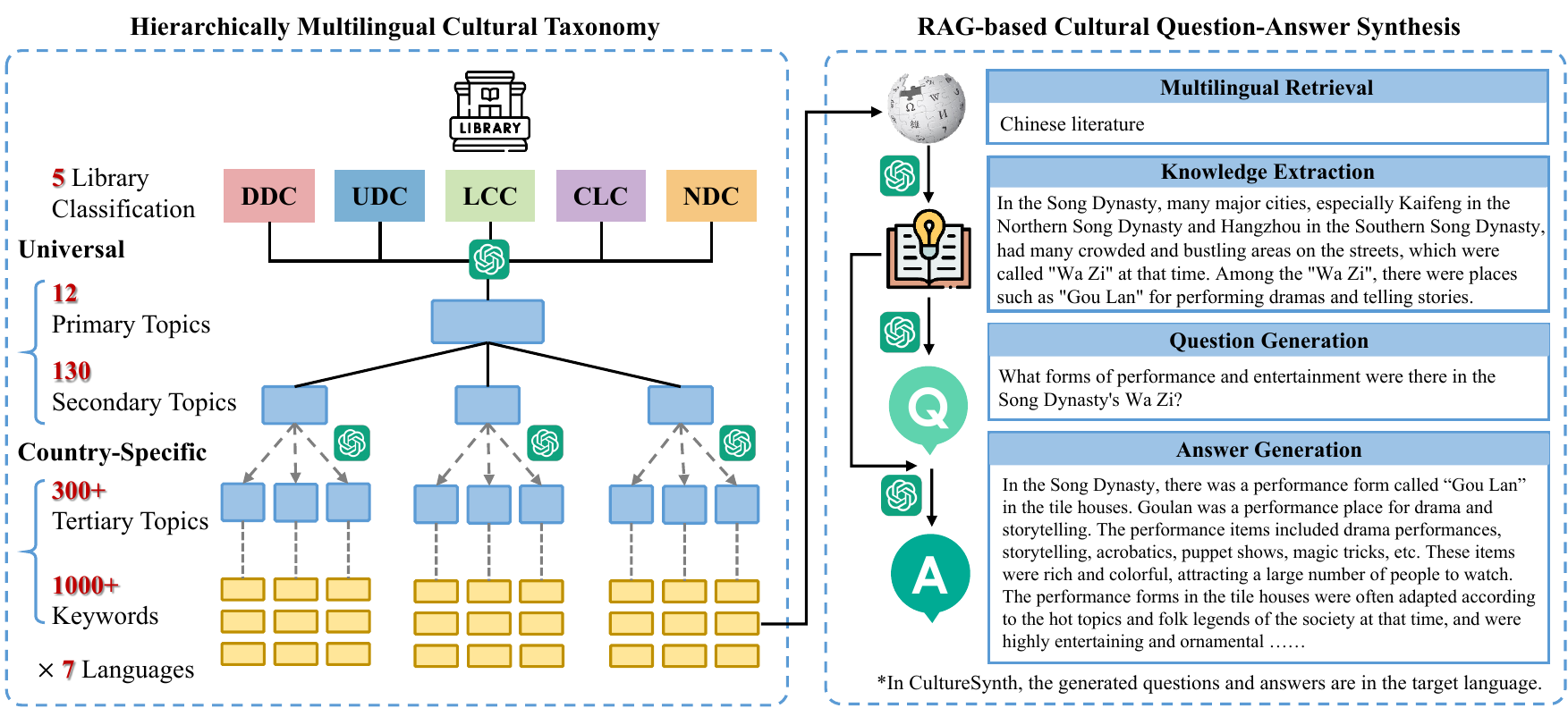}
  \vspace{-4mm}
  \caption{Overview of CultureSynth: (1) a hierarchically multilingual cultural taxonomy (see Section~\ref{sec:culturalframework}), and (2) an RAG-based question-answer synthesis methodology (see Section~\ref{sec:datagen}). The cultural taxonomy integrates five library classifications from different cultural backgrounds to establish primary topics, which are then hierarchically expanded using LLMs to create comprehensive and detailed cultural tertiary topics and keywords. The synthesis method leverages cultural keywords to automatically perform multilingual knowledge retrieval, extract factual knowledge, and generate question-answer pairs through LLMs.}
  \label{fig:framework}
  \vspace{-4mm}
\end{figure*}

Existing cultural competence benchmarks for LLMs primarily fall into two categories: multiple-choice questions (MCQ) and question-answer generation \cite{pawar2024survey,minaee2024large}. In the MCQ category, MMLU \cite{hendrycksmeasuring} covers 57 subjects across humanities, social sciences, and STEM, while BLEnD \cite{myung2024blend} evaluates everyday cultural knowledge across diverse languages. CulturalBench \cite{chiu2024culturalbench} features 1,227 human-verified questions assessing cultural knowledge across 45 regions and 17 topics. Additionally, NORMAD \cite{rao2024normad} specifically analyzes LLMs' behavior under varying socio-cultural contexts.

Question-answer generation benchmarks better evaluate cultural behavioral alignment and interaction styles. CALMQA \cite{arora2024calmqa} collects questions from community forums across 12 topics with native speaker validation, while NativQA \cite{hasan2024nativqa} employs Google's "People also ask" feature to generate queries across 18 topics efficiently. CultureBank \cite{shi2024culturebank} provides 23,000 structured cultural data entries from social media platforms using 11 predefined fields, enabling more flexible interpretation. PRISM \cite{kirk2024prism} contributes preference data from diverse annotators across 23 topic clusters. While these efforts advance cultural competence evaluation, our work focuses on expanding domain coverage and reducing data generation costs through automated synthesis methods.

\subsection{Synthesizing Data for LLMs}
\vspace{-1mm}


The growth of LLMs has increased demands for high-quality training data, particularly in culturally sensitive domains where multilingual resources are scarce. Synthetic data generation has emerged as a promising solution for addressing scalability and domain-specific adaptability~\cite{liu2024best}. Recent studies by \citet{gan2024towards} show that synthetic data can enhance model generalization through post-training information gain.
However, concerns about synthetic data reliability persist due to LLM hallucinations \citep{DBLP:journals/csur/JiLFYSXIBMF23}. RAG frameworks have effectively addressed this by anchoring outputs in verified knowledge bases \citep{DBLP:journals/corr/abs-2404-10981}, improving answer faithfulness by over 30\% \citep{DBLP:journals/corr/abs-2402-19473}. Additionally, role-playing prompts and instruction tuning have proven effective in refining synthetic outputs, with \citet{DBLP:journals/corr/abs-2410-14251} demonstrating improved performance through multi-agent systems.
Our work builds on these advances by combining hierarchical cultural taxonomies with retrieval-augmented synthesis to produce culturally nuanced and verifiable data.

\section{CultureSynth}


This section presents the complete question-answer generation process of the CultureSynth framework, as illustrated in Figure~\ref{fig:framework}.

\subsection{Hierarchically Cultural Taxonomy} \label{sec:culturalframework}

The taxonomy comprises two tiers: universal cultural topics and country-specific cultural topics. The first tier establishes cross-country analytical topics through standardized cultural dimensions, while the second tier incorporates localized cultural elements corresponding to specific linguistic contexts. This hierarchical structure offers dual advantages: the universal tier ensures comprehensive analysis of cultural competence by preventing analytical redundancy or omissions caused by cross-cultural discrepancies, whereas the country-specific tier preserves linguistic and cultural nuances essential for granular cultural interpretation.

\vspace{1mm}
\noindent\textbf{Universal Cultural Topics.}
We integrate five multinational library classification systems, i.e., Dewey Decimal Classification (DDC), Universal Decimal Classification (UDC), Library of Congress Classification (LCC), Chinese Library Classification (CLC), and Nippon Decimal Classification (NDC), to establish a cultural framework comprising primary and secondary topics. This cross-cultural synthesis ensures comprehensive coverage of universal cultural dimensions across diverse national contexts. Then through a combination of LLM-powered analysis and expert curation, we ultimately derived 12 primary topics: Social Sciences (SS), Philosophy and Psychology (PP), Religion and Theology (RT), Political Science (PS), Law (LAW), Education (EDU), Language (LAN), Literature (LIT), Medicine (MED), Applied Sciences and Technology (AST), Arts (ART), and Recreation, Sports, and Entertainment (RSE), along with 130 secondary topics, as listed in Table~\ref{tab:topics_all}.

\vspace{1mm}
\noindent\textbf{Country-Specific Cultural Topics.}
Based on primary and secondary topics, we designed a role-playing-based prompt to guide the LLM (e.g., we use GPT-4~\cite{achiam2023gpt}) in generating more fine-grained cultural tertiary topics and keywords for each country (or language), as shown in Figure~\ref{fig:prompt1}. We observe that augmenting the prompt with the instruction "\textit{Please don't be lazy and answer this question in depth from a local's perspective}" leads to more comprehensive and culturally localized responses.
Ultimately, our approach facilitates the construction of more than $300$ fine-grained tertiary topics, with an average of over $1000$ keywords per country/ language\footnote{Tertiary topics and keywords are available at \url{https://github.com/Eyr3/CultureSynth}.}.



\subsection{RAG-based Multilingual Cultural Question-Answer synthesis} \label{sec:datagen}

The method consists of four main steps: First, we perform multilingual retrieval to obtain authentic and reliable information pages corresponding to the given keywords, followed by extracting key knowledge points. Based on these verified knowledge points, we then automatically generate high-quality questions and answers. Throughout the generation process, prompt optimization is applied to ensure the logical coherence of the questions and the comprehensiveness and depth of the answers. 

\vspace{1mm}
\noindent\textbf{Step1: Multilingual Retrieval.} 
Given a target keyword, we first translate it into country-specific languages and retrieve corresponding pages from reliable sources (e.g., Wikipedia) in both English and the target language. We employ LLMs with prompts detailed in Figures~\ref{fig:prompt2} and \ref{fig:prompt3} to assess the retrieved pages for keyword relevance and culture-specific content. Pages failing to meet keyword relevance or cultural content requirements are excluded from subsequent knowledge extraction.





\vspace{1mm}
\noindent\textbf{Step2: Knowledge Extraction.}
For verified culturally significant pages, we employ a prompt detailed in Figures~\ref{fig:prompt4} and \ref{fig:prompt5} that enables LLMs to systematically extract multiple knowledge points using a standardized JSON format. This stage ensures structural consistency while preserving cultural specificity in the extracted information.




\vspace{1mm}
\noindent\textbf{Step3: Question Generation.}
For each knowledge point, we use LLMs with the prompt in Figures~\ref{fig:prompt6} and \ref{fig:prompt7} to generate questions in the target language, following three key guidelines: avoiding offensive or discriminatory content, eliminating anaphoric references to ensure context clarity, and maintaining cultural and linguistic appropriateness.






\vspace{1mm}
\noindent\textbf{Step4: Answer Generation.}
To construct comprehensive answers, we prompt the LLM to act as a domain expert, instructing it to incorporate cultural knowledge and provide detailed responses in the target language that address multiple aspects of each question (see Figures~\ref{fig:prompt8}). Similar to questions, answers must avoid demonstrative pronouns to maintain contextual independence. 





\section{CultureSynth-7 Benchmark}
\begin{table*}[!t]
\centering
\footnotesize
\caption{The acceptance rates (\%) demonstrate inter-annotator consistency and high data quality across 7 languages.}
\vspace{-2mm}
\resizebox{\linewidth}{!}{
\begin{tabular}{lccccccccc}
\toprule
 & Annotator & ar & zh & fr & ja & ko & pt & es & Average \\
 \midrule
\multirow{3}{*}{\begin{tabular}[c]{@{}l@{}}Question Clarity Rate \\ (score: 1) \end{tabular}} & Annotator A & 85.8 & 98.3 & 85.8 & 95.0 & 98.3 & 85.0 & 93.3 & 91.6 \\
 & Annotator B & 100.0 & 99.2 & 100.0 & 100.0 & 100.0 & 100.0 & 100.0 & 99.9 \\
 & Average & 92.9 & 98.8 & 92.9 & 97.5 & 99.2 & 92.5 & 96.7 & 95.8 \\
  \midrule
\multirow{3}{*}{\begin{tabular}[c]{@{}l@{}}Cultural Relevance Rate \\ (score: 1)\end{tabular}} & Annotator A & 72.8 & 92.4 & 87.6 & 90.4 & 67.8 & 92.2 & 98.2 & 85.9 \\
 & Annotator B & 70.0 & 92.4 & 75.0 & 87.5 & 66.7 & 90.0 & 91.7 & 81.9 \\
 & Average & 71.4 & 92.4 & 81.3 & 89.0 & 67.3 & 91.1 & 95.0 & 83.9 \\
  \midrule
\multirow{3}{*}{\begin{tabular}[c]{@{}l@{}}High-quality Answer Rate \\ (score: $\geq 4$)\end{tabular}} & Annotator A & 100.0 & 95.8 & 99.0 & 98.2 & 100.0 & 99.0 & 97.3 & 98.5 \\
 & Annotator B & 100.0 & 100.0 & 98.3 & 97.5 & 100.0 & 99.2 & 98.3 & 99.0 \\
 & Average & 100.0 & 97.9 & 98.7 & 97.9 & 100.0 & 99.1 & 97.8 & 98.8 \\
\bottomrule
\end{tabular}
\vspace{-8mm}
\label{tab:manual_results}
}
\vspace{-4mm}
\end{table*}

\subsection{Data Annotation}
To validate the quality of synthetic data in CultureSynth-7, we randomly sample 120 QA pairs per language. The assessment involves two native speakers (annotators  A and B) assigned to each language, totaling 14 annotators. To ensure annotation consistency and reliability, we implement a comprehensive training protocol that includes guideline alignment and trial annotation.

\vspace{1mm}
\noindent\textbf{Annotation Guideline.}
This guideline outlines the criteria for assessing QA pairs through three dimensions (refer to Appendix~\ref{sec:q_score} for details).
\vspace{-2mm}
\begin{itemize}[leftmargin=*]
\item{Question Clarity \& Safety:} Determine if the question is self-contained and adheres to universal ethical standards (Score: 1 for yes, 0 for no).
\vspace{-2mm}
\item{Cultural Relevance:} Identify cultural distinctiveness based on two dimensions. Score 1 if the question shows either cultural variance (answers differ across cultures/languages) or cultural specificity (contains culture-specific elements such as regional traditions); score 0 otherwise.
\vspace{-2mm}
\item{Answer Quality:} Access the quality of the answers relative to the reference knowledge (i.e., Wikipedia) using the 5-point scale, with scores of $\geq 4$ being high quality.
\end{itemize}
\vspace{-2mm}

\vspace{1mm}
\noindent\textbf{Annotation Results.} 
Table~\ref{tab:manual_results} presents the acceptance rates of three assessment dimensions across 7 languages. 
In the cultural relevance assessment, most languages show minimal differences between annotators, with some languages achieving high agreement (e.g., zh and ja). Similarly, for the high-quality answer rate, the variations between annotators are typically less than 2\%, with several languages showing strong consistency (e.g., ar and ko both at 100\%). While we observe annotation variations of 14.2\% to 15\% in question clarity rate for ar, fr, and pt, these discrepancies can be primarily attributed to different interpretations of question self-consistency between annotators. Nevertheless, it's noteworthy that the lowest question clarity rate still achieves 85\% for these three languages. 

Regarding data quality, all three dimensions show strong performance across languages. The question clarity rate achieves a high overall average of 95.41\%, with particularly strong results in zh and ko. The cultural relevance rate, while slightly lower, maintains a robust average of 83.91\%, with es and zh performing notably well. Notably, the high-quality answer rate remains consistently strong across all languages, averaging 98.76\%, with ar and ko reaching 100\% and other languages scoring above 97\%. These consistently high scores across all three metrics strongly validate the effectiveness of our data construction approach.


Regarding safety considerations, our question safety assessment in our annotation reveals negligible safety concerns in the generated questions, primarily due to two factors: the use of Wikipedia as the source material, which inherently limits safety risks, and the employment of LLMs with built-in safe mechanisms for question generation.

\subsection{Data Release and Analysis}
\vspace{-1mm}


From over 300 tertiary topics per language in Section~\ref{sec:culturalframework}, we generate 19,360 QA pairs, split into two disjoint subsets with non-overlapping tertiary topics: a 4,149-example CultureSynth-7-mini\footnote{We use \textit{CultureSynth-7} to identify \textit{CultureSynth-7-mini}, which are manually verified and released publicly.} set for model development and resource-constrained users, and a 15,211-example CultureSynth-7-max for additional evaluation or model fine-tuning. To maintain data integrity, CultureSynth-7-max sets remain private, and can be made available upon request via email. To achieve representative sampling, we randomly selected 20 tertiary topics from each primary topic category for CultureSynth-7, including all associated QA pairs. This approach ensures zero overlaps in tertiary topics between CultureSynth-7 and CultureSynth-7-max, effectively preventing data contamination. 

\begin{figure*}[!t]
    \begin{minipage}[t]{0.48\textwidth}
        \centering
        \includegraphics[width=\linewidth]{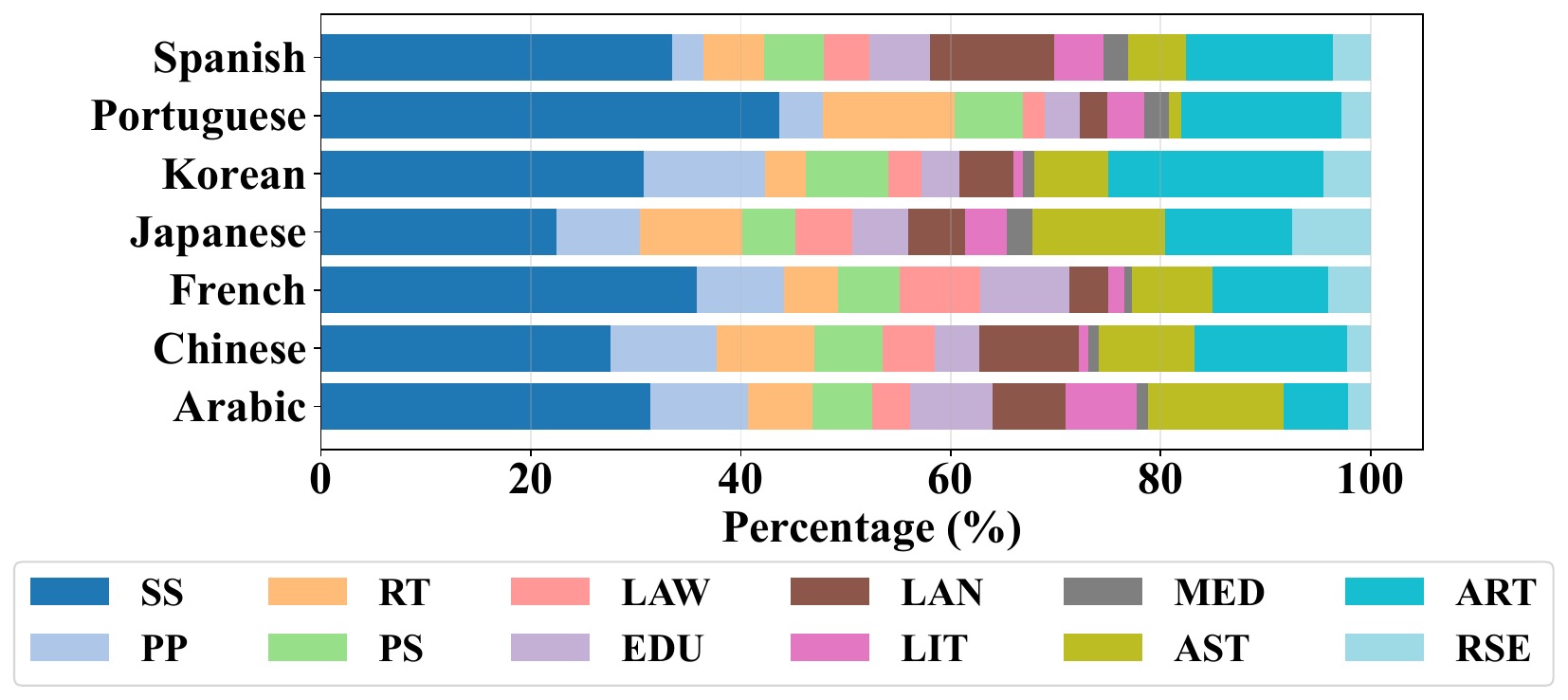}
        \vspace{-6mm}
        \caption{Topic distribution for total 19,360 QA pairs.}
        \label{fig:topic_distribution_all}
    \end{minipage}
    \hfill
    \begin{minipage}[t]{0.48\textwidth}
        \centering
        \includegraphics[width=\linewidth]{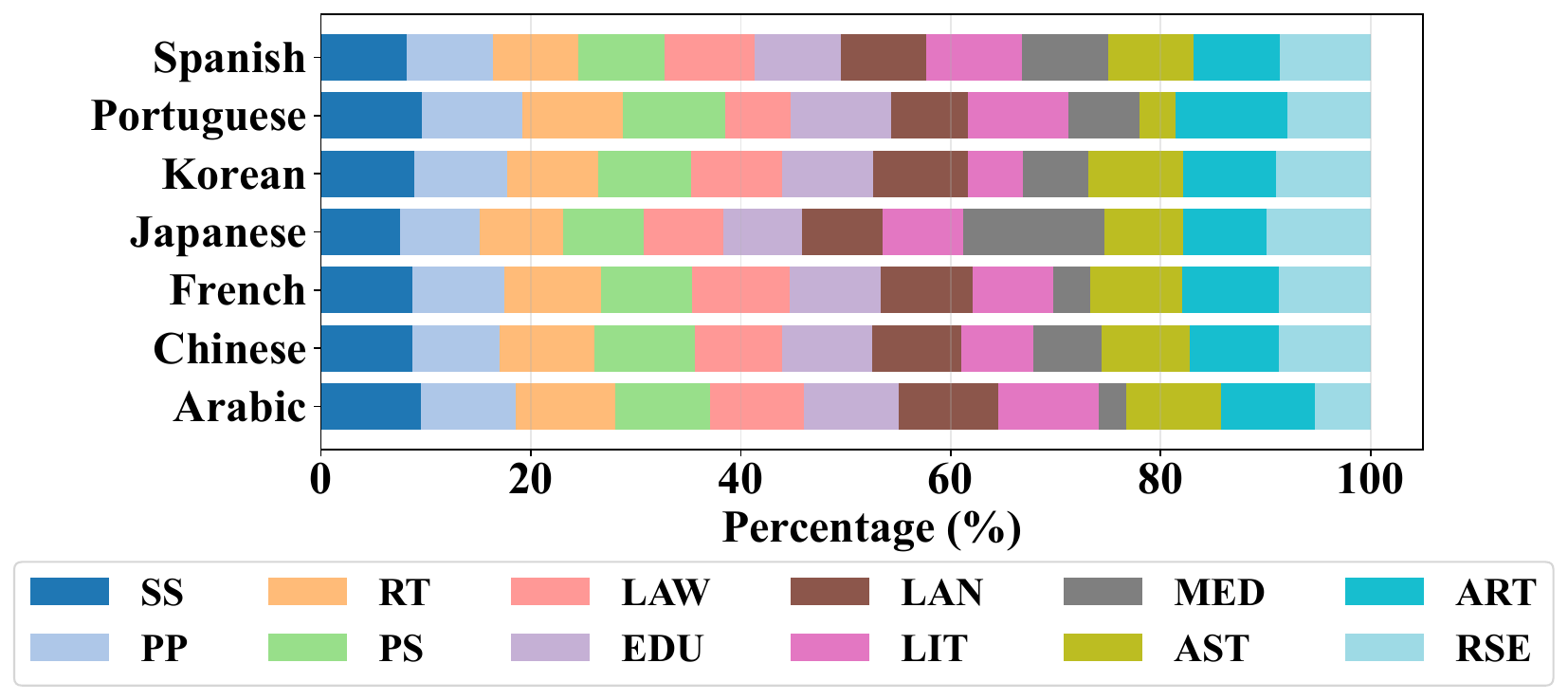}
        \vspace{-6mm}
        \caption{Balanced topic distribution in CultureSynth-7.}
        \label{fig:topic_distribution_mini}
    \end{minipage}
    \vspace{-3mm}
\end{figure*}

\begin{table*}[ht]
\centering
\footnotesize
\caption{Key statistics of the total 19,360 QA pairs in 7 languages from different language families. }
\vspace{-2mm}
\begin{tabular}{cccccccccc}
\toprule
\multicolumn{1}{c}{\multirow{2}{*}{Language}} &
  \multirow{2}{*}{Code} &
  \multicolumn{1}{c}{\multirow{2}{*}{Languag family}} &
  \multirow{2}{*}{Total QA} &
  \multirow{2}{*}{Percentage} &
  \multicolumn{2}{c}{Question length} &
  \multicolumn{3}{c}{Answer length} \\
\cmidrule(r){6-7} \cmidrule(r){8-10}
\multicolumn{1}{c}{} &    & \multicolumn{1}{c}{} &      &        & Avg & Max & Avg & Max & Std \\
\midrule                                   
Spanish              & es & Indo-European        & 2170 & 11.2\% & 26  & 77  & 90  & 458 & 110 \\
French               & fr & Indo-European        & 2929 & 15.1\% & 25  & 67  & 66  & 417 & 90  \\
Portuguese           & pt & Indo-European        & 1529 & 7.9\%  & 25  & 77  & 98  & 451 & 110 \\
Arabic               & ar & Afro-Asiatic         & 1416 & 7.3\%  & 19  & 52  & 87  & 310 & 88  \\
Chinese              & zh & Sino-Tibetan         & 4172 & 21.5\% & 22  & 86  & 62  & 560 & 102 \\
Japanese             & ja & Japonic              & 3798 & 19.6\% & 29  & 112 & 77  & 516 & 109 \\
Korean               & ko & Koreanic             & 3346 & 17.3\% & 15  & 51  & 38  & 310 & 56   \\
\bottomrule
\end{tabular}
\label{tab:statistics}
\vspace{-4mm}
\end{table*}

Table~\ref{tab:statistics} presents statistics for the total of 19,360 QA pairs. Question lengths average 15 and 29 words, with language-specific variations, while answer lengths range from 38-98 words on average, reaching 560 words in Chinese. Figure~\ref{fig:topic_distribution_all} shows cultural topic distribution of these QA pairs, revealing distinct patterns across languages, with Arabic emphasizing social sciences and Chinese focusing on arts and applied sciences. For evaluation, we use the balanced and manually verified CultureSynth-7 benchmark (containing 50 to 54 samples per language and topic, see Figure~\ref{fig:topic_distribution_mini}), with all questions annotated for clarity and cultural relevance, and answers verified for high quality.

\section{Experiments}

\subsection{Experiment Protocol}
\vspace{-1mm}

Given the subjective nature of CultureSynth and its substantial size (4,149 questions), we employ an LLM-based automatic evaluation approach~\cite{zheng2023judging,liu2023g,chiangchatbot}. This involves a pairwise comparison using a high-performance LLM as the judge LLM. For each question in CultureSynth, we obtain responses from a moderate-performance baseline model and a target model. The judge LLM then compares both responses against the reference answer, assigning scores as follows: $1$ if the target model performs better, $-1$ if the baseline model performs better, and $0$ if performances are comparable. As referenced in \cite{chiangchatbot}, the specific prompt is shown in Figure~\ref{fig:prompt9}. After completing this process for all questions, we calculate the net win rate of the target model against the baseline.

\begin{figure*}[!t]
    \centering
    \begin{subfigure}{0.4\textwidth}
        \includegraphics[width=\textwidth]{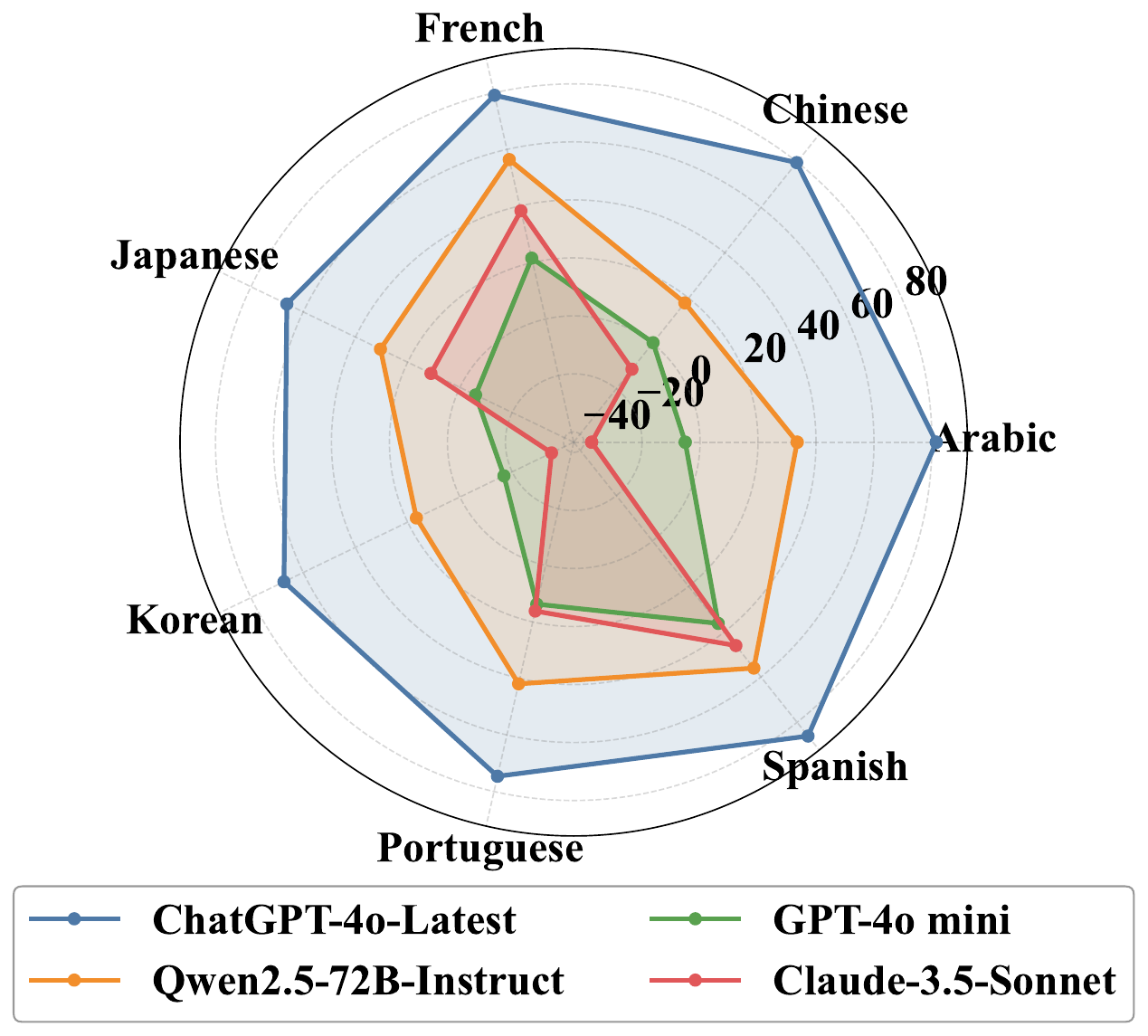}
        \vspace{-6mm}
        \caption{First-tier cultural competence models}
        \label{fig:rate_lang_1}
    \end{subfigure}
    \begin{subfigure}{0.4\textwidth}
        \includegraphics[width=\textwidth]{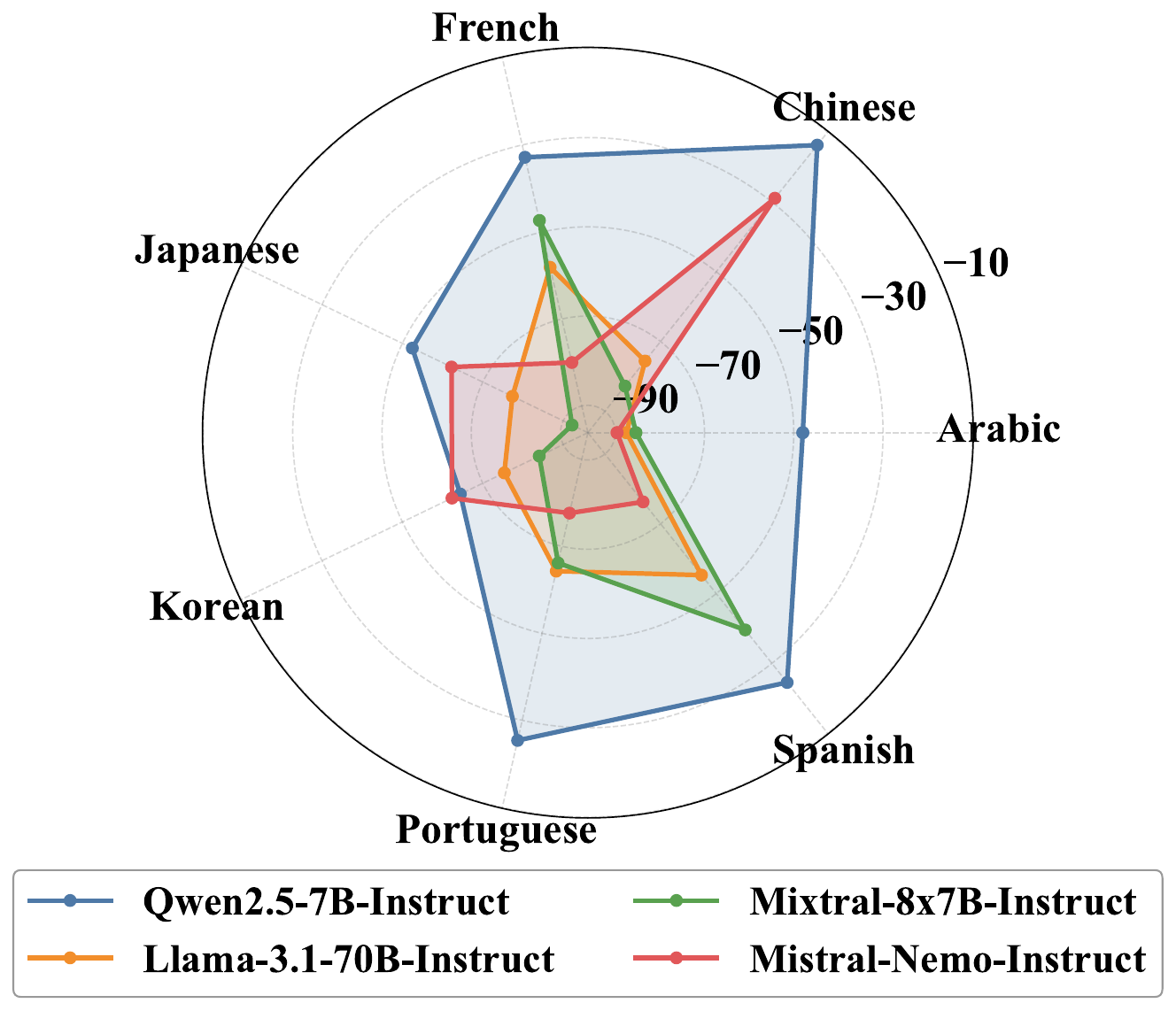}
        \vspace{-6mm}
        \caption{Second-tier cultural competence models}
        \label{fig:rate_lang_2}
    \end{subfigure}
    \vspace{-2mm}
    \caption{Net win rates (\%) of different models compared to the baseline model across languages.}
    \vspace{-3mm}
\end{figure*}

\begin{table*}[!t]
\centering
\setlength\tabcolsep{5pt}
\footnotesize
\caption{Net win rates (\%) of models across languages, relative to the Gemma2-27B-it baseline under the ChatGPT-4o-Latest judge. {\color{blue}Blue} and {\color{yellow}yellow} indicate the best and second-best performing models, respectively.}
\vspace{-2mm}
\resizebox{\linewidth}{!}{
\begin{tabular}{lcccccccc}
\toprule
{\color[HTML]{1F2329} Model}                      & Arabic       & Chinese       & French       & Japanese       & Korean       & Portuguese       & Spanish       & Weighted Average \\
\midrule
{\color[HTML]{1F2329} ChatGPT-4o-Latest} & \cellcolor[HTML]{E1EAFF}{\color[HTML]{1F2329} 81.48} & \cellcolor[HTML]{E1EAFF}{\color[HTML]{1F2329} 79.77} & \cellcolor[HTML]{E1EAFF}{\color[HTML]{1F2329} 79.15} & \cellcolor[HTML]{E1EAFF}{\color[HTML]{1F2329} 66.27} & \cellcolor[HTML]{E1EAFF}{\color[HTML]{1F2329} 67.36} & \cellcolor[HTML]{E1EAFF}{\color[HTML]{1F2329} 74.62} & \cellcolor[HTML]{E1EAFF}{\color[HTML]{1F2329} 86.06} & \cellcolor[HTML]{E1EAFF}{\color[HTML]{1F2329} 76.31} \\
{\color[HTML]{1F2329} GPT-4o mini} & {\color[HTML]{1F2329} -5.11} & {\color[HTML]{1F2329} 0.33} & {\color[HTML]{1F2329} 21.54} & {\color[HTML]{1F2329} -6.00} & {\color[HTML]{1F2329} -16.84} & {\color[HTML]{1F2329} 13.72} & {\color[HTML]{1F2329} 36.38} & {\color[HTML]{1F2329} 6.31} \\
\midrule
{\color[HTML]{1F2329} Claude-3.5-Sonnet} & {\color[HTML]{1F2329} -37.39} & {\color[HTML]{1F2329} -11.37} & {\color[HTML]{1F2329} 38.29} & {\color[HTML]{1F2329} 11.09} & {\color[HTML]{1F2329} -35.07} & {\color[HTML]{1F2329} 16.17} & {\color[HTML]{1F2329} 46.15} & {\color[HTML]{1F2329} 4.58} \\
\midrule
{\color[HTML]{1F2329} Qwen2.5-72B-Instruct} & \cellcolor[HTML]{FAF1D1}{\color[HTML]{1F2329} 33.51} & \cellcolor[HTML]{FAF1D1}{\color[HTML]{1F2329} 17.89} & \cellcolor[HTML]{FAF1D1}{\color[HTML]{1F2329} 56.41} & \cellcolor[HTML]{FAF1D1}{\color[HTML]{1F2329} 30.43} & \cellcolor[HTML]{FAF1D1}{\color[HTML]{1F2329} 16.67} & \cellcolor[HTML]{FAF1D1}{\color[HTML]{1F2329} 41.92} & \cellcolor[HTML]{FAF1D1}{\color[HTML]{1F2329} 56.09} & \cellcolor[HTML]{FAF1D1}{\color[HTML]{1F2329} 36.13} \\
{\color[HTML]{1F2329} Qwen2.5-7B-Instruct} & {\color[HTML]{1F2329} -47.97} & {\color[HTML]{1F2329} -13.71} & {\color[HTML]{1F2329} -32.82} & {\color[HTML]{1F2329} -52.47} & {\color[HTML]{1F2329} -64.41} & {\color[HTML]{1F2329} -25.38} & {\color[HTML]{1F2329} -24.52} & {\color[HTML]{1F2329} -37.48} \\
{\color[HTML]{1F2329} Qwen2.5-3B-Instruct} & {\color[HTML]{1F2329} -86.77} & {\color[HTML]{1F2329} -32.27} & {\color[HTML]{1F2329} -83.59} & {\color[HTML]{1F2329} -87.41} & {\color[HTML]{1F2329} -83.16} & {\color[HTML]{1F2329} -74.81} & {\color[HTML]{1F2329} -73.08} & {\color[HTML]{1F2329} -74.48} \\
{\color[HTML]{1F2329} Qwen2.5-0.5B-Instruct} & {\color[HTML]{1F2329} -99.82} & {\color[HTML]{1F2329} -84.62} & {\color[HTML]{1F2329} -97.44} & {\color[HTML]{1F2329} -97.45} & {\color[HTML]{1F2329} -96.70} & {\color[HTML]{1F2329} -96.43} & {\color[HTML]{1F2329} -96.47} & {\color[HTML]{1F2329} -95.54} \\
\midrule
{\color[HTML]{1F2329} Llama-3.1-70B-Instruct} & {\color[HTML]{1F2329} -87.48} & {\color[HTML]{1F2329} -75.59} & {\color[HTML]{1F2329} -58.12} & {\color[HTML]{1F2329} -77.36} & {\color[HTML]{1F2329} -75.35} & {\color[HTML]{1F2329} -64.29} & {\color[HTML]{1F2329} -55.29} & {\color[HTML]{1F2329} -70.50} \\
{\color[HTML]{1F2329} Llama-3.1-8B-Instruct} & {\color[HTML]{1F2329} -92.77} & {\color[HTML]{1F2329} -89.13} & {\color[HTML]{1F2329} -73.33} & {\color[HTML]{1F2329} -90.85} & {\color[HTML]{1F2329} -90.62} & {\color[HTML]{1F2329} -74.81} & {\color[HTML]{1F2329} -75.96} & {\color[HTML]{1F2329} -84.07} \\
{\color[HTML]{1F2329} Llama-3.2-3B-Instruct} & {\color[HTML]{1F2329} -99.82} & {\color[HTML]{1F2329} -95.32} & {\color[HTML]{1F2329} -97.09} & {\color[HTML]{1F2329} -98.20} & {\color[HTML]{1F2329} -98.44} & {\color[HTML]{1F2329} -90.41} & {\color[HTML]{1F2329} -93.11} & {\color[HTML]{1F2329} -96.12} \\
{\color[HTML]{1F2329} Llama-3.2-1B-Instruct} & {\color[HTML]{1F2329} -99.47} & {\color[HTML]{1F2329} -93.48} & {\color[HTML]{1F2329} -96.75} & {\color[HTML]{1F2329} -98.20} & {\color[HTML]{1F2329} -98.26} & {\color[HTML]{1F2329} -94.36} & {\color[HTML]{1F2329} -94.39} & {\color[HTML]{1F2329} -96.43} \\
\midrule
{\color[HTML]{1F2329} Mixtral-8x7B-Instruct-v0.1} & {\color[HTML]{1F2329} -85.36} & {\color[HTML]{1F2329} -82.78} & {\color[HTML]{1F2329} -47.35} & {\color[HTML]{1F2329} -92.20} & {\color[HTML]{1F2329} -84.03} & {\color[HTML]{1F2329} -66.17} & {\color[HTML]{1F2329} -39.58} & {\color[HTML]{1F2329} -71.20} \\
{\color[HTML]{1F2329} Mistral-Nemo-Instruct-2407} & {\color[HTML]{1F2329} -89.59} & {\color[HTML]{1F2329} -28.93} & {\color[HTML]{1F2329} -80.00} & {\color[HTML]{1F2329} -62.22} & {\color[HTML]{1F2329} -62.33} & {\color[HTML]{1F2329} -77.63} & {\color[HTML]{1F2329} -76.28} & {\color[HTML]{1F2329} -67.77} \\
{\color[HTML]{1F2329} Mistral-7B-Instruct-v0.3} & {\color[HTML]{1F2329} -98.77} & {\color[HTML]{1F2329} -88.96} & {\color[HTML]{1F2329} -86.15} & {\color[HTML]{1F2329} -95.20} & {\color[HTML]{1F2329} -94.27} & {\color[HTML]{1F2329} -81.20} & {\color[HTML]{1F2329} -83.97} & {\color[HTML]{1F2329} -89.90}
\\ \bottomrule
\end{tabular}
}
\label{tab:net_win_rate_lang}
\vspace{-4mm}
\end{table*}

\subsection{Experimental Setup}
\vspace{-1mm}

We select ChatGPT-4o-Latest (version gpt-4o-2024-05-13)~\cite{achiam2023gpt} and Qwen3‑32B‑Think~\cite{yang2025qwen3} as our judge LLMs, Gemma2-27B-it~\cite{team2024gemma} as our baseline model, and then evaluate the following target models: GPT-4o series~\cite{achiam2023gpt},
Claude-3.5-Sonnet~\cite{anthropic2024claude35}, Qwen2.5 series~\cite{yang2024qwen2},
Llama 3 series~\cite{dubey2024llama}, 
and Mixtral series~\cite{jiang2024mixtral}. 
Responses of proprietary models are obtained through their respective APIs. 
%

Our evaluation metric is the net win rate:
\vspace{-2mm}
\[
\scalebox{0.95}{$
\text{Net win rate} = {(N_{\text{target\_wins}} - N_{\text{baseline\_wins}})}/{N_{\text{total}}},$}
\]
where $N_{\text{target\_wins}}$ denotes the number of cases where the score is $1$ (indicating the target model is superior), $N_{\text{baseline\_wins}}$ represents the number of cases with a score of $-1$ (indicating the baseline model is superior), and $N_{\text{total}}$ is the total number of comparisons.

\subsection{Experimental Results} \label{sec:results}
\vspace{-1mm}

Based on the evaluation results across seven languages and 12 primary topics presented in Table~\ref{tab:net_win_rate_lang}, our CultureSynth benchmark reveals clear performance stratification among different models. The overall ranking of cultural competence follows this order: ChatGPT-4o-Latest $>$ Qwen2.5-72B-Instruct $>$ GPT-4o mini $\approx$ Claude-3.5-Sonnet $>$ Qwen2.5-7B-Instruct $>$  Mistral-Nemo-Instruct-2407 $\approx$ Llama-3.1-70B-Instruct $\approx$ Mixtral-8x7B-Instruct-v0.1 $>$ Qwen2.5-3B-Instruct $>$ Llama-3.1-8B-Instruct $>$ Mistral-7B-Instruct-v0.3 $>$ Qwen2.5-0.5B-Instruc $\approx$ Llama-3.2-3B-Instruct $\approx$ Llama-3.2-1B-Instruct. 




\vspace{1mm}
\noindent\textbf{Cross-lingual Performance Analysis.}
As illustrated in Figure~\ref{fig:rate_lang_1}), among first-tier models, ChatGPT-4o-Latest excels across most languages but shows limitations in East Asian cultural contexts (Japanese/ Korean). GPT-4o mini exhibits similar language performance patterns to ChatGPT-4o-Latest. Qwen2.5-72B-Instruct demonstrates strong generalization in medium-resource languages like Portuguese and Japanese. Claude-3.5-Sonnet performs well in French and Spanish but struggles with Arabic and Korean. 
Second-tier models, though performing below baseline, show distinct language strengths (as shown in Figure~\ref{fig:rate_lang_2}): Qwen2.5-7B-Instruct excels in Chinese, while Llama-3.1-70B-Instruct and Mixtral-8x7B-Instruct perform better in French and Spanish. Mistral-Nemo-Instruct shows strength in Chinese compared to its performance in other languages.

As shown in Figure~\ref{fig:rate_lang_size}, model cultural competence correLatest positively with size across all languages. Models below 3B parameters experience a significant drop in language capabilities (see Figure~\ref{fig:rate_lang_all_2}), only managing to handle culture-related QA in specific native language scenarios.

\begin{figure*}[!t]
    \centering
    \begin{subfigure}{0.4\textwidth}
        \includegraphics[width=\textwidth]{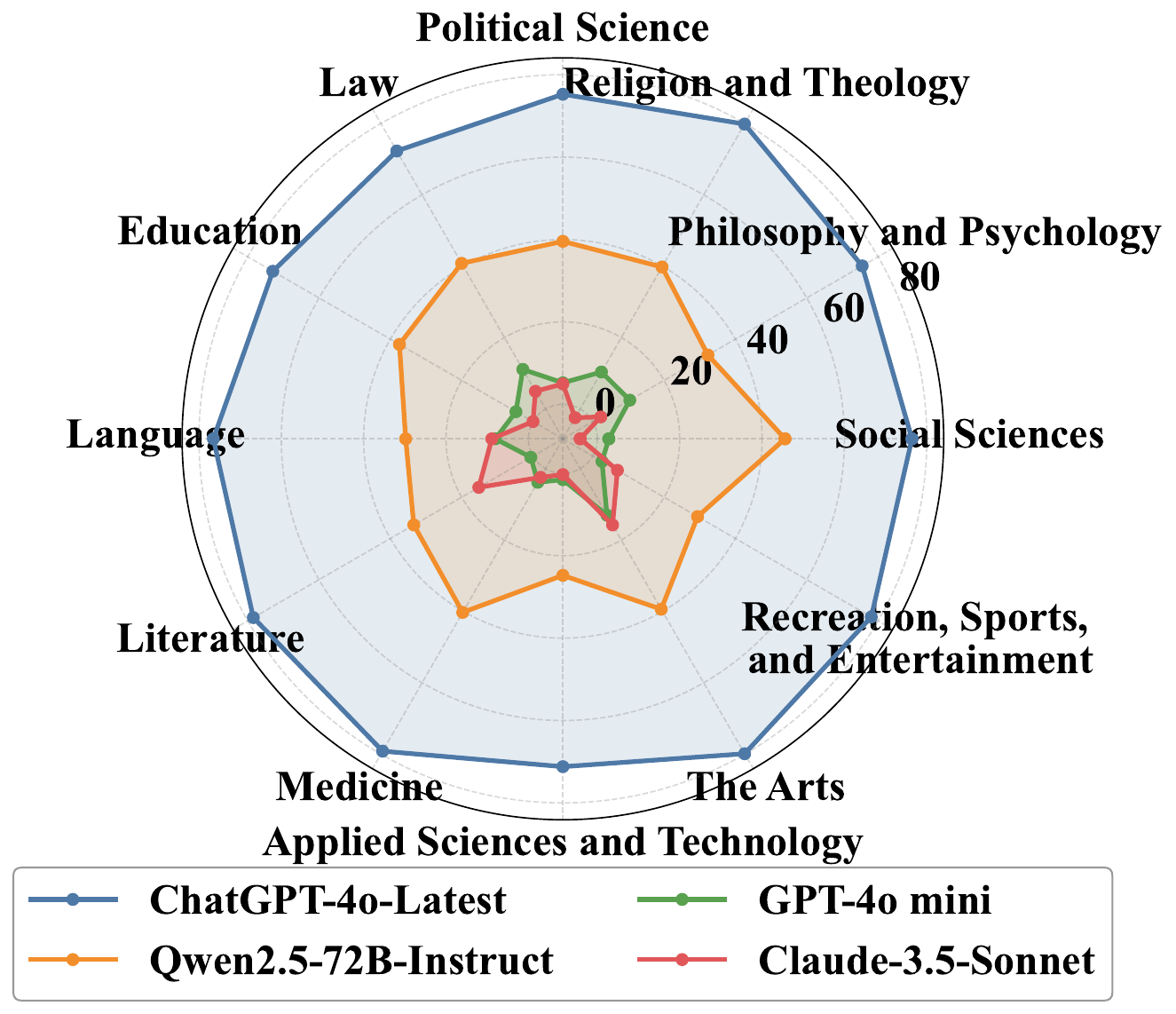}
        \vspace{-6mm}
        \caption{First-tier cultural competence models}
        \label{fig:rate_topic_1}
    \end{subfigure}
    \begin{subfigure}{0.4\textwidth}
        \includegraphics[width=\textwidth]{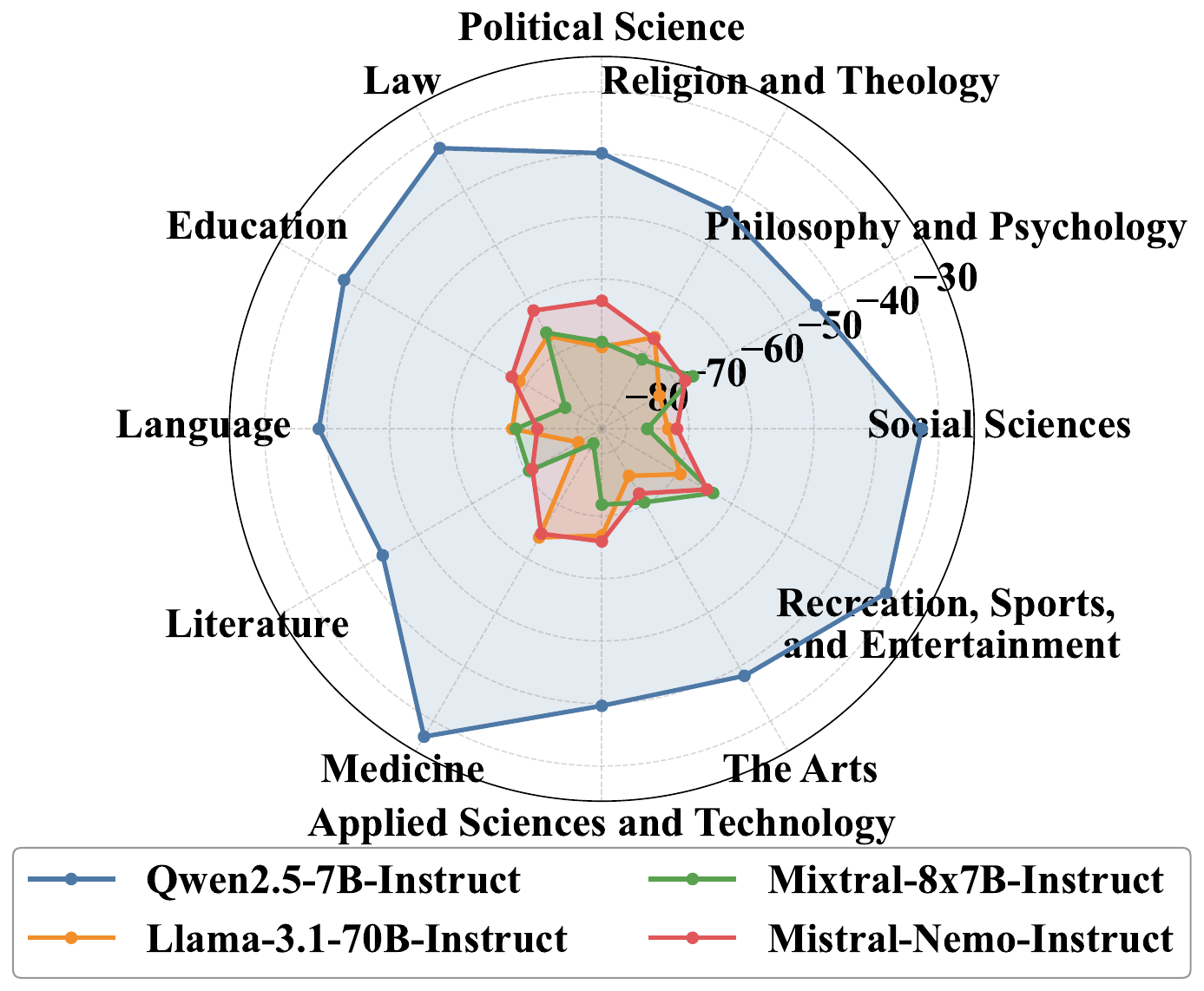}
        \vspace{-6mm}
        \caption{Second-tier cultural competence models}
        \label{fig:rate_topic_2}
    \end{subfigure}
    
    \vspace{-2mm}
    \caption{Net win rates (\%) of different models compared to the baseline model across cultural topics.}
    \vspace{-3mm}
\end{figure*}

\begin{table*}[!t]
\footnotesize
\setlength\tabcolsep{2pt}
\caption{Net win rates (\%) of models across cultural topics, relative to the Gemma2-27B-it baseline under the ChatGPT-4o-Latest judge. {\color{blue}Blue} and {\color{yellow}yellow} indicate the best and second-best performing models, respectively.}
\vspace{-2mm}
\resizebox{\linewidth}{!}{
\begin{tabular}{lcccccccccccc}
\toprule  
Model & \begin{tabular}[c]{@{}c@{}}Social \\ Sciences\end{tabular} & \begin{tabular}[c]{@{}c@{}}Philosophy \\ and \\ Psychology\end{tabular} & \begin{tabular}[c]{@{}c@{}}Religion \\ and \\ Theology\end{tabular} & \begin{tabular}[c]{@{}c@{}}Political \\ Science\end{tabular} & Law & \begin{tabular}[c]{@{}c@{}}Edu-\\cation\end{tabular} & \begin{tabular}[c]{@{}c@{}}Lang-\\uage\end{tabular} & \begin{tabular}[c]{@{}c@{}}Liter-\\ature\end{tabular} & \begin{tabular}[c]{@{}c@{}}Medi-\\cine\end{tabular} & \begin{tabular}[c]{@{}c@{}}Applied \\ Sciences and \\ Technology\end{tabular} & \begin{tabular}[c]{@{}c@{}}Arts\end{tabular}   & \begin{tabular}[c]{@{}c@{}}Recreation, \\ Sports, and \\ Entertainment\end{tabular} \\
\midrule
{\color[HTML]{1F2329} ChatGPT-4o-Latest} & \cellcolor[HTML]{E1EAFF}76.39 & \cellcolor[HTML]{E1EAFF}75.56 & \cellcolor[HTML]{E1EAFF}79.84 & \cellcolor[HTML]{E1EAFF}75.27 & \cellcolor[HTML]{E1EAFF}72.30 & \cellcolor[HTML]{E1EAFF}72.96 & \cellcolor[HTML]{E1EAFF}76.50 & \cellcolor[HTML]{E1EAFF}78.42 & \cellcolor[HTML]{E1EAFF}79.17 & \cellcolor[HTML]{E1EAFF}71.21 & \cellcolor[HTML]{E1EAFF}79.89 & \cellcolor[HTML]{E1EAFF}78.10 \\
{\color[HTML]{1F2329} GPT-4o mini} & 2.78 & 10.39 & 10.35 & 5.22 & 11.08 & 4.79 & 8.02 & 0.61 & 3.82 & 1.55 & 13.04 & 2.59 \\
\midrule
{\color[HTML]{1F2329} Claude-3.5-Sonnet} & -4.17 & 2.25 & -2.45 & 4.95 & 4.96 & 0.00 & 8.88 & 15.20 & 2.43 & 0.31 & 15.76 & 6.92 \\
\midrule
{\color[HTML]{1F2329} Qwen2.5-72B-Instruct} & \cellcolor[HTML]{FAF1D1}45.56 & \cellcolor[HTML]{FAF1D1}32.30 & \cellcolor[HTML]{FAF1D1}39.78 & \cellcolor[HTML]{FAF1D1}39.56 & \cellcolor[HTML]{FAF1D1}40.82 & \cellcolor[HTML]{FAF1D1}37.46 & \cellcolor[HTML]{FAF1D1}29.80 & \cellcolor[HTML]{FAF1D1}33.43 & \cellcolor[HTML]{FAF1D1}40.28 & \cellcolor[HTML]{FAF1D1}24.77 & \cellcolor[HTML]{FAF1D1}39.40 & \cellcolor[HTML]{FAF1D1}29.39 \\
{\color[HTML]{1F2329} Qwen2.5-7B-Instruct} & -32.78 & -44.38 & -43.87 & -39.84 & -32.07 & -36.34 & -38.68 & -43.47 & -27.08 & -39.63 & -38.32 & -31.41 \\
{\color[HTML]{1F2329} Qwen2.5-3B-Instruct} & -68.89 & -76.69 & -80.11 & -75.27 & -67.64 & -76.90 & -73.64 & -81.76 & -75.69 & -70.28 & -71.74 & -75.22 \\
{\color[HTML]{1F2329} Qwen2.5-0.5B-Instruct} & -98.33 & -96.35 & -97.28 & -96.43 & -93.59 & -96.34 & -95.13 & -96.66 & -94.10 & -96.90 & -94.02 & -91.07 \\
\midrule
{\color[HTML]{1F2329} Llama-3.1-70B-Instruct} & -73.33 & -73.31 & -67.03 & -70.88 & -66.76 & -68.73 & -69.63 & -79.64 & -63.89 & -66.87 & -75.27 & -69.45 \\
{\color[HTML]{1F2329} Llama-3.1-8B-Instruct} & -85.56 & -82.30 & -88.01 & -82.42 & -84.55 & -82.82 & -85.67 & -86.63 & -81.25 & -85.45 & -80.71 & -83.29 \\
{\color[HTML]{1F2329} Llama-3.2-3B-Instruct} & -97.22 & -96.91 & -96.73 & -95.05 & -94.17 & -98.03 & -96.85 & -97.87 & -95.83 & -98.45 & -92.66 & -93.95 \\
{\color[HTML]{1F2329} Llama-3.2-1B-Instruct} & -97.50 & -96.63 & -96.46 & -93.13 & -96.79 & -97.75 & -97.71 & -97.26 & -96.53 & -97.52 & -95.65 & -94.52 \\
\midrule
{\color[HTML]{1F2329} Mixtral-8x7B-Instruct-v0.1} & -76.67 & -67.13 & -71.12 & -70.05 & -66.18 & -77.18 & -70.20 & -70.52 & -81.25 & -71.83 & -70.38 & -63.40 \\
{\color[HTML]{1F2329} Mistral-Nemo-Instruct-2407} & -71.94 & -68.54 & -67.30 & -63.46 & -62.10 & -67.32 & -73.64 & -71.12 & -64.58 & -65.94 & -72.01 & -64.55 \\
{\color[HTML]{1F2329} Mistral-7B-Instruct-v0.3} & -90.83 & -92.70 & -88.83 & -90.93 & -88.05 & -93.52 & -91.12 & -92.10 & -88.19 & -88.85 & -85.60 & -87.90 \\
\midrule
\rowcolor[HTML]{DEE0E3} 
Average & -48.04 & -48.17 & -47.80 & -46.60 & -44.48 & -48.55 & -47.79 & -49.24 & -45.91 & -48.85 & -44.88 & -45.55
\\
\bottomrule
\end{tabular}
}
\label{tab:net_win_rate_topic}
\vspace{-4mm}
\end{table*}

\vspace{1mm}
\noindent\textbf{Topic-wise Performance Analysis.}
According to Table~\ref{tab:net_win_rate_topic}, different models exhibit significant variations in their cultural topic comprehension capabilities. Cultural competence is not solely determined by the total number of parameters but is closely related to the organization of domain knowledge, the weighting of cultural data during training, and the model architectures.

For first-tier models (as shown in Figure~\ref{fig:rate_topic_1}), GPT-4o-Latest demonstrates exceptional cultural literacy, particularly in topics requiring deep cultural context understanding (e.g., literature and medicine). Qwen2.5-72B-Instruct achieves 50-60\% of GPT-4o-Latest's performance in practical domains like social sciences, law, and medicine, but shows relative weakness in creative fields such as language and arts. Its capability profile indicates stronger performance in structured knowledge applications, reflecting its training emphasis on professional content. Claude-3.5-Sonnet displays uneven cultural competence distribution, showing certain advantages in generative tasks (e.g., literature and arts) while demonstrating weaknesses in domains requiring objective knowledge and reasoning (e.g., social sciences and religious theology).

\begin{table*}[!t]
\footnotesize
\setlength\tabcolsep{3pt}
\caption{Net win rates (\%) of ChatGPT-4o-Latest across languages and cultural topics ({\color{red}red}: below 60\%).}
\vspace{-2mm}
\resizebox{\linewidth}{!}{
\begin{tabular}{lcccccccccccc}
\toprule  
& \begin{tabular}[c]{@{}c@{}}Social \\ Sciences\end{tabular} & \begin{tabular}[c]{@{}c@{}}Philosophy \\ and \\ Psychology\end{tabular} & \begin{tabular}[c]{@{}c@{}}Religion \\ and \\ Theology\end{tabular} & \begin{tabular}[c]{@{}c@{}}Political \\ Science\end{tabular} & Law & \begin{tabular}[c]{@{}c@{}}Education\end{tabular} & \begin{tabular}[c]{@{}c@{}}Language\end{tabular} & \begin{tabular}[c]{@{}c@{}}Literature\end{tabular} & \begin{tabular}[c]{@{}c@{}}Medicine\end{tabular} & \begin{tabular}[c]{@{}c@{}}Applied \\ Sciences and \\ Technology\end{tabular} & \begin{tabular}[c]{@{}c@{}}Arts\end{tabular}   & \begin{tabular}[c]{@{}c@{}}Recreation, \\ Sports, and \\ Entertainment\end{tabular} \\
\midrule
ar     & 83.33 & 76.47 & 85.19 & 88.24 & 78.43 & 82.35 & 88.89 & 81.48 & 73.33 & 76.47 & 80.39 & 73.33 \\
zh    & 88.46 & 86.00 & 79.63 & 84.21 & 64.00 & 76.47 & 74.51 & 90.24 & 76.92 & 72.00 & 82.35 & 82.69 \\
fr     & 72.55 & 76.47 & 74.07 & 64.71 & 74.07 & 84.31 & 84.31 & 80.00 & 90.48 & 88.24 & 81.48 & 86.27 \\
ja   & 78.00 & {\color{red}52.94} & 73.58 & 68.63 & 64.71 & {\color{red}54.00} & 64.71 & 62.75 & 71.11 & 64.00 & 77.36 & 60.61 \\
ko     & 68.63 & 68.63 & 76.00 & 62.75 & 60.00 & {\color{red}56.00} & {\color{red}59.62} & 73.33 & 75.00 & {\color{red}53.85} & 82.35 & 76.92 \\
es    & 80.39 & 92.16 & 84.31 & 84.31 & 87.04 & 82.35 & 88.24 & 84.21 & 98.04 & 76.47 & 80.39 & 94.44 \\
pt & 62.75 & 76.47 & 86.27 & 73.08 & 78.79 & 74.51 & 74.36 & 76.47 & 75.00 & 61.11 & 75.44 & 73.81
\\
\bottomrule
\end{tabular}
}
\label{tab:net_win_rate_model_1}
\vspace{-3mm}
\end{table*}

\begin{table*}[!t]
\footnotesize
\setlength\tabcolsep{3pt}
\caption{Net win rates (\%) of GPT-4o mini across languages and cultural topics ({\color{red}red}: below -20\%).}
\vspace{-2mm}
\resizebox{\linewidth}{!}{
\begin{tabular}{lcccccccccccc}
\toprule  
& \begin{tabular}[c]{@{}c@{}}Social \\ Sciences\end{tabular} & \begin{tabular}[c]{@{}c@{}}Philosophy \\ and \\ Psychology\end{tabular} & \begin{tabular}[c]{@{}c@{}}Religion \\ and \\ Theology\end{tabular} & \begin{tabular}[c]{@{}c@{}}Political \\ Science\end{tabular} & Law & \begin{tabular}[c]{@{}c@{}}Education\end{tabular} & \begin{tabular}[c]{@{}c@{}}Language\end{tabular} & \begin{tabular}[c]{@{}c@{}}Literature\end{tabular} & \begin{tabular}[c]{@{}c@{}}Medicine\end{tabular} & \begin{tabular}[c]{@{}c@{}}Applied \\ Sciences and \\ Technology\end{tabular} & \begin{tabular}[c]{@{}c@{}}Arts\end{tabular}   & \begin{tabular}[c]{@{}c@{}}Recreation, \\ Sports, and \\ Entertainment\end{tabular} \\
\midrule
ar & 3.70   & {-15.69} & 1.85   & 15.69  & 13.73 & -3.92  & 0.00   & {\color{red} -46.30} & 6.67   & -7.84  & -3.92 & {\color{red}-23.33} \\
zh & 7.69   & 16.00  & {\color{red}-20.37} & 1.75   & -6.00 & -9.80  & 0.00   & 19.51  & 5.13   & -6.00  & 7.84  & -5.77  \\
fr & 19.61  & 45.10  & 29.63  & 7.84   & 18.52 & 29.41  & 15.69  & 4.44   & 33.33  & 35.29  & 27.78 & -3.92  \\
ja & {\color{red}-26.00} & {-13.73} & -7.55  & 3.92   & 7.84  & 2.00   & {-17.65} & {-17.65} & {-12.22} & -10.00 & 18.87 & 1.52   \\
ko & {\color{red}-27.45} & {-11.76} & 0.00   & {\color{red}-33.33} & 2.00  & {\color{red}-24.00} & -3.85  & -6.67  & {\color{red}-33.33} & {\color{red}-30.77} & -7.84 & {\color{red}-25.00} \\
pt & 5.88   & 15.69  & 33.33  & 13.46  & 15.15 & -9.80  & -5.13  & 15.69  & 16.67  & 5.56   & 14.04 & 40.48  \\
es & 35.29  & 37.25  & 37.25  & 27.45  & 25.93 & 49.02  & 64.71  & 35.09  & 35.29  & 27.45  & 33.33 & 29.63 
\\
\bottomrule
\end{tabular}
}
\label{tab:net_win_rate_model_2}
\vspace{-4mm}
\end{table*}

For second-tier models (as shown in Figure~\ref{fig:rate_topic_2}), Qwen2.5-7B-Instruct and Llama-3.1-70B-Instruct perform slightly better in professional domains like medicine compared to other topics. Qwen2.5-7B-Instruct maintains baseline capabilities in highly structured fields like medicine and law. However, these models show significant deficiencies in humanities, such as philosophy, psychology, and literature, with performance declining by over 40\%. Llama-3.1-70B-Instruct's superior performance in medicine reveals its strength in Western knowledge systems, while its poor performance in the humanities exposes cultural adaptation deficiencies. Comparing results between Mixtral-8x7B-Instruct and Mistral-Nemo-Instruct reveals that the mixture-of-experts architecture demonstrates effective routing for discrete knowledge points, whereas the dense transformer with a longer context window shows higher performance in political science and law domains requiring long-range textual dependencies.

For extremely small-scale models in the third and fourth-tiers (as shown in Figure~\ref{fig:rate_topic_all_2}), cultural comprehension capabilities essentially collapse when parameters are reduced to the 1B level.

\vspace{1mm}
\noindent\textbf{Model-wise Cultural Competence Analysis.}
We conduct a model-wise analysis of cultural competence for two representative models, the best-performing ChatGPT-4o-Latest and its smaller counterpart GPT-4o mini, to illustrate performance variation across cultural and linguistic contexts.

Table~\ref{tab:net_win_rate_model_1} shows that ChatGPT‑4o‑Latest delivers strong and balanced performance across languages and topics, with notable strengths in Spanish, Arabic, and Chinese contexts. However, its relatively lower scores in Philosophy and Psychology (ja) and Applied Sciences and Technology (ko) highlight persistent challenges in certain language–domain pairs.
In contrast, Table~\ref{tab:net_win_rate_model_2} indicates that GPT-4o mini excels in Indo-European languages, especially within humanities and entertainment topics, but underperforms in Arabic literature and several Korean and Japanese domains. These disparities underscore the need for targeted improvements to enhance its cultural competence.


\vspace{1mm}
\noindent\textbf{Consistency of Results Across Judge Models.}
Table~\ref{tab:net_win_rate_lang_1} reports the language-specific net win rates under the Qwen3‑32B‑Think judge. Comparing these results with those in Table~\ref{tab:net_win_rate_lang} (ChatGPT-4o-Latest judge) shows that, while absolute values vary slightly, the overall model ranking by weighted average net win rate is preserved:
ChatGPT-4o-Latest $>$ Qwen2.5-72B-Instruct $>$ Claude-3.5-Sonnet $\approx$ GPT-4o mini $>$ Qwen2.5-7B-Instruct $>$ Llama-3.1-70B-Instruct $\approx$ Mistral-Nemo-Instruct-2407 $\approx$ Mixtral-8x7B-Instruct-v0.1 $>$ remaining models.
Language-specific performance trends are also consistent. For instance, {Qwen2.5-72B-Instruct} excels in French and Spanish regardless of the judge model.
%
Similarly, Table~\ref{tab:net_win_rate_topic_1} reports the cultural topic-wise net win rates under the Qwen3‑32B‑Think judge. Comparison with Table~\ref{tab:net_win_rate_topic} (ChatGPT-4o-Latest judge) reveals parallel topic-specific patterns. For example, {Claude-3.5-Sonnet} consistently ranks highest in {arts} and relatively lower in {religion and theology} under both judges.


We further compute the agreement rate (the proportion of evaluation instances where both judges yield identical pairwise preferences) between ChatGPT-4o-Latest and Qwen3‑32B‑Think judgments for each model. As shown in Table~\ref{tab:agreement_rate}, the average agreement rates across all 14 models reaches 85.05\%. 

These findings indicate that LLM-based evaluations of cultural competence, when using a capable judge model and comprehensive prompts, are robust and largely judge-independent.



\section{Conclusion}
\vspace{-2mm}
In this paper, we present CultureSynth, a framework for synthesizing culturally-aware QA pairs to evaluate LLMs' cultural competence. By combining a hierarchically cultural taxonomy with RAG-based synthesis, we construct CultureSynth-7, demonstrating high-quality synthetic data generation. Our extensive evaluation, spanning 14 LLMs across 7 languages and 12 cultural topics, reveals that cultural competence depends on multiple factors beyond model scale, including training data composition, architectural design, and knowledge organization. We also identify different cultural understanding gaps, particularly in geographic and domain-specific contexts.

\section*{Limitations}
\vspace{-2mm}
Our study has two main limitations. First, the uneven distribution of keywords across languages and topics creates inherent dataset imbalances in the total $19,360$ QA pairs, despite our balanced sampling approach in the evaluation benchmark (i.e., CultureSynth-7). Second, when analyzing cultural topics, we did not categorize questions based on their cognitive demands (e.g., creative thinking, factual knowledge) and difficulty levels, which provides an opportunity for future refinement through fine-grained question taxonomies.


\bibliography{custom}

\appendix

\section{Human Annotation}

\subsection{Detailed Annotation Criteria} \label{sec:q_score}

\noindent\textbf{Question Clarity \& Safety:} Determine if the question is self-contained and adheres to universal ethical standards. 

\begin{itemize}[nosep]  
    \item Score 1: Question is clear, comprehensible, and self-contained
    \item Score 0: Question exhibits any of the following issues: requires additional context for comprehension, contains demonstrative pronouns without context, or contains unsafe elements (violence, explicit content, hate speech).
    
\end{itemize}

\noindent\textbf{Cultural Relevance:} Identify cultural distinctiveness through dual dimensions.

\begin{itemize}[nosep]
    \item Score 1: Question that exhibits either cultural variance (answers differ across cultures/languages) or cultural specificity (containing culture-specific elements such as regional traditions). 
    \item Score 0: Question lacks cultural elements or specificity.
\end{itemize}

\noindent\textbf{Answer Quality:} Access the quality of the answers relative to the reference knowledge (i.e., Wikipedia) using the 5-point scale, with scores of $\geq$ 4 being high quality.
\begin{itemize}[nosep]
\item Score 5: Exceptional answer (comprehensive, accurate, well-structured)
\item Score 4: Strong answer (minor improvements possible)
\item Score 3: Adequate answer (notable omissions or inaccuracies)
\item Score 2: Insufficient answer (major gaps or errors)
\item Score 1: Poor answer (significantly flawed)
\item Score 0: Unacceptable answer (incorrect or inappropriate)
\end{itemize}



\subsection{Characteristics Of Annotators}

Our annotation process maintains rigorous professional standards throughout. For each of the seven languages (ar, es, fr, ja, ko, pt, and zh), two native speakers were recruited from a professional annotation service provider as annotators to ensure quality. The entire annotation task was completed within a two-week period. The annotation cost varied by language, ranging from approximately 0.57 to 1.71 USD for each QA pair.

\section{Additional Experimental Results}
\label{sec:appendix}
\vspace{-1mm}

Table~\ref{tab:agreement_rate} shows the average agreement rates across all 14 models.
Table~\ref{tab:net_win_rate_lang_1} and \ref{tab:net_win_rate_topic_1} report the language-specific and cultural topic-wise net win rates under the Qwen3-32B-Think judge, respectively.

\begin{figure}[!t]
    \centering
    \begin{subfigure}{0.45\textwidth}
        \includegraphics[width=\textwidth]{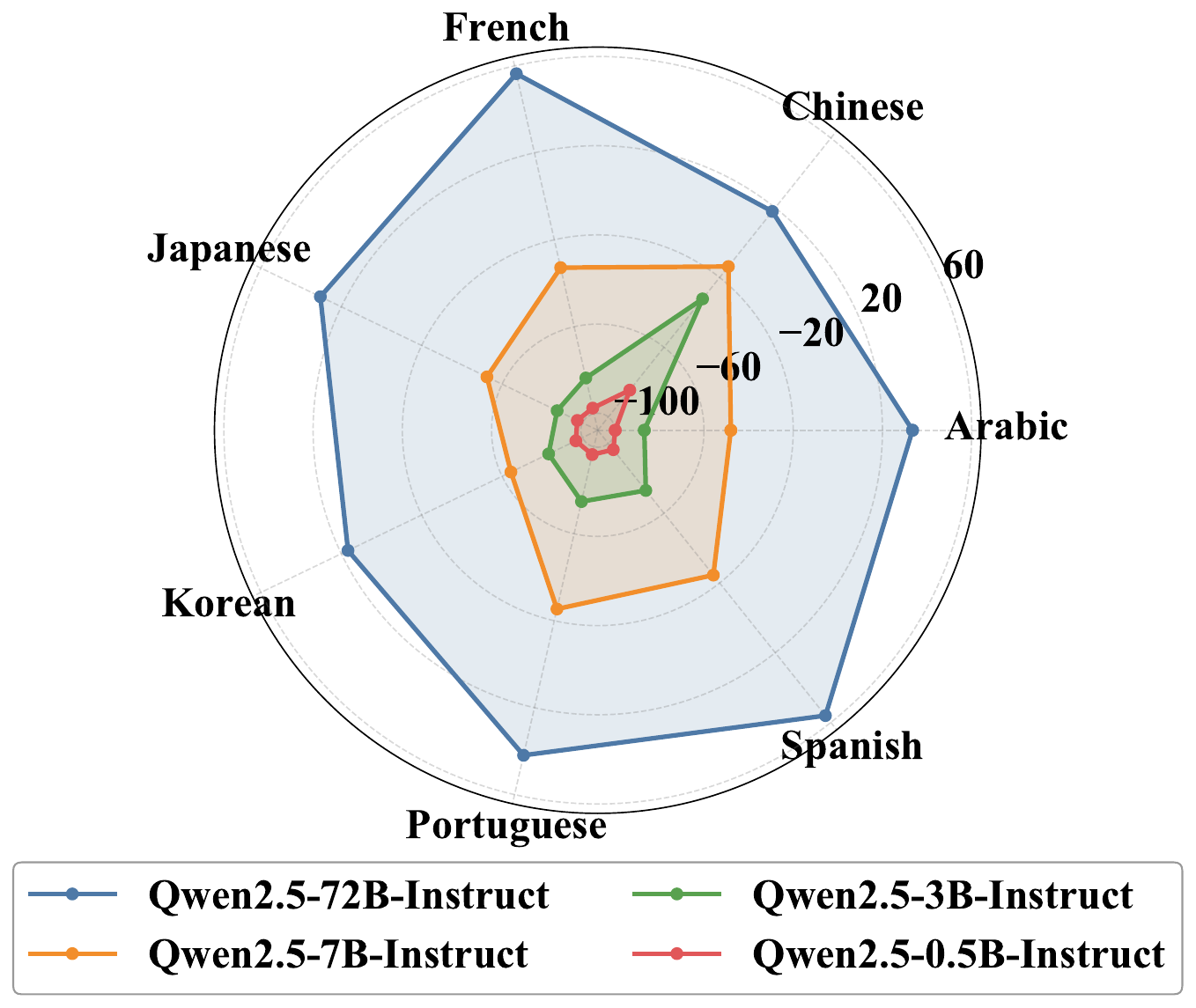}
        \caption{Qwen2.5 series}
    \end{subfigure}
    \begin{subfigure}{0.45\textwidth}
        \includegraphics[width=\textwidth]{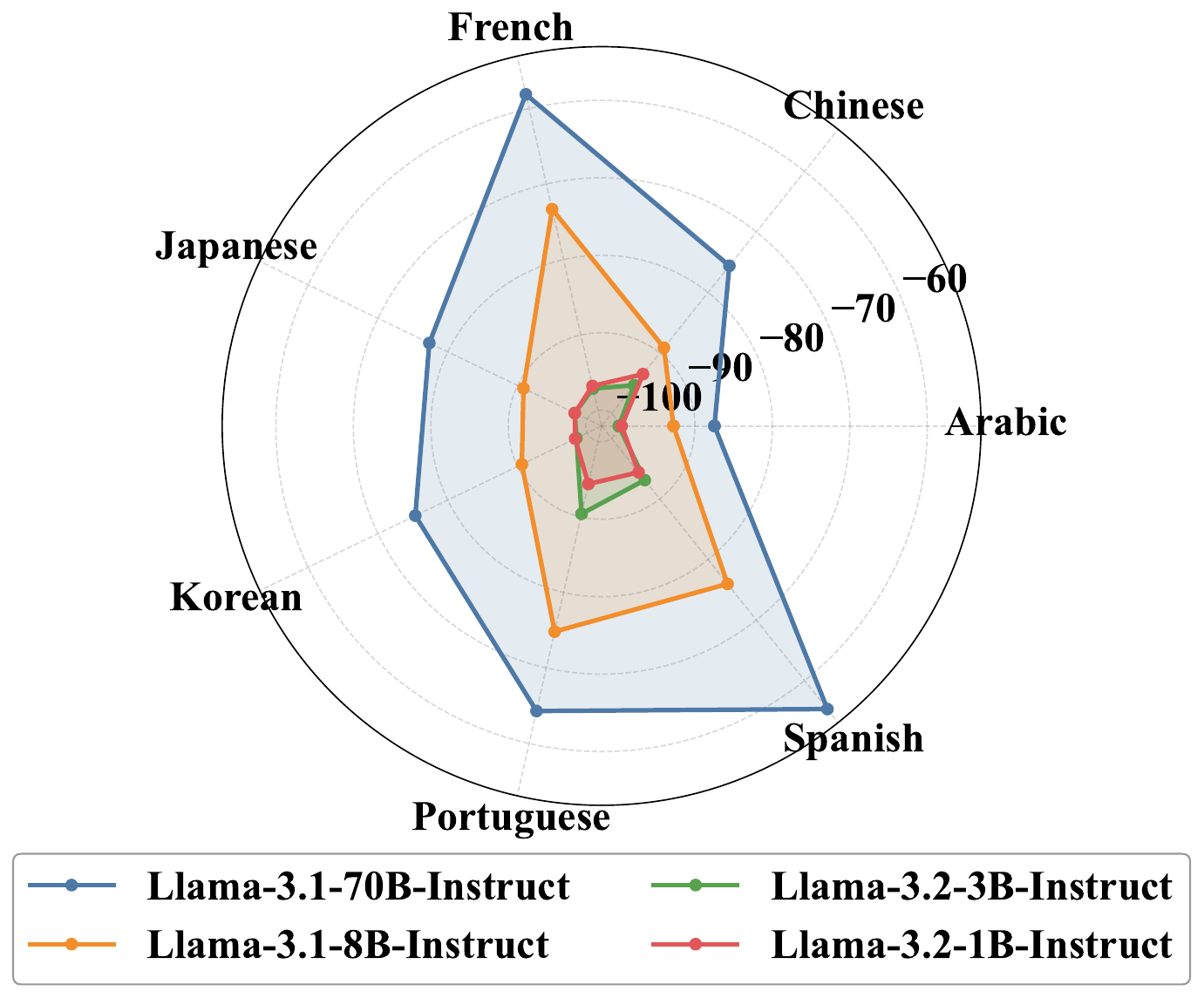}
        \caption{Llama 3 series}
    \end{subfigure}
    \vspace{-3mm}
    \caption{Net win rates (\%) of models of different sizes compared to the baseline model across languages.}
    \vspace{-3mm}
    \label{fig:rate_lang_size}
\end{figure}

\begin{figure*}[!t]
    \centering
    \begin{subfigure}{0.45\textwidth}
        \includegraphics[width=\textwidth]{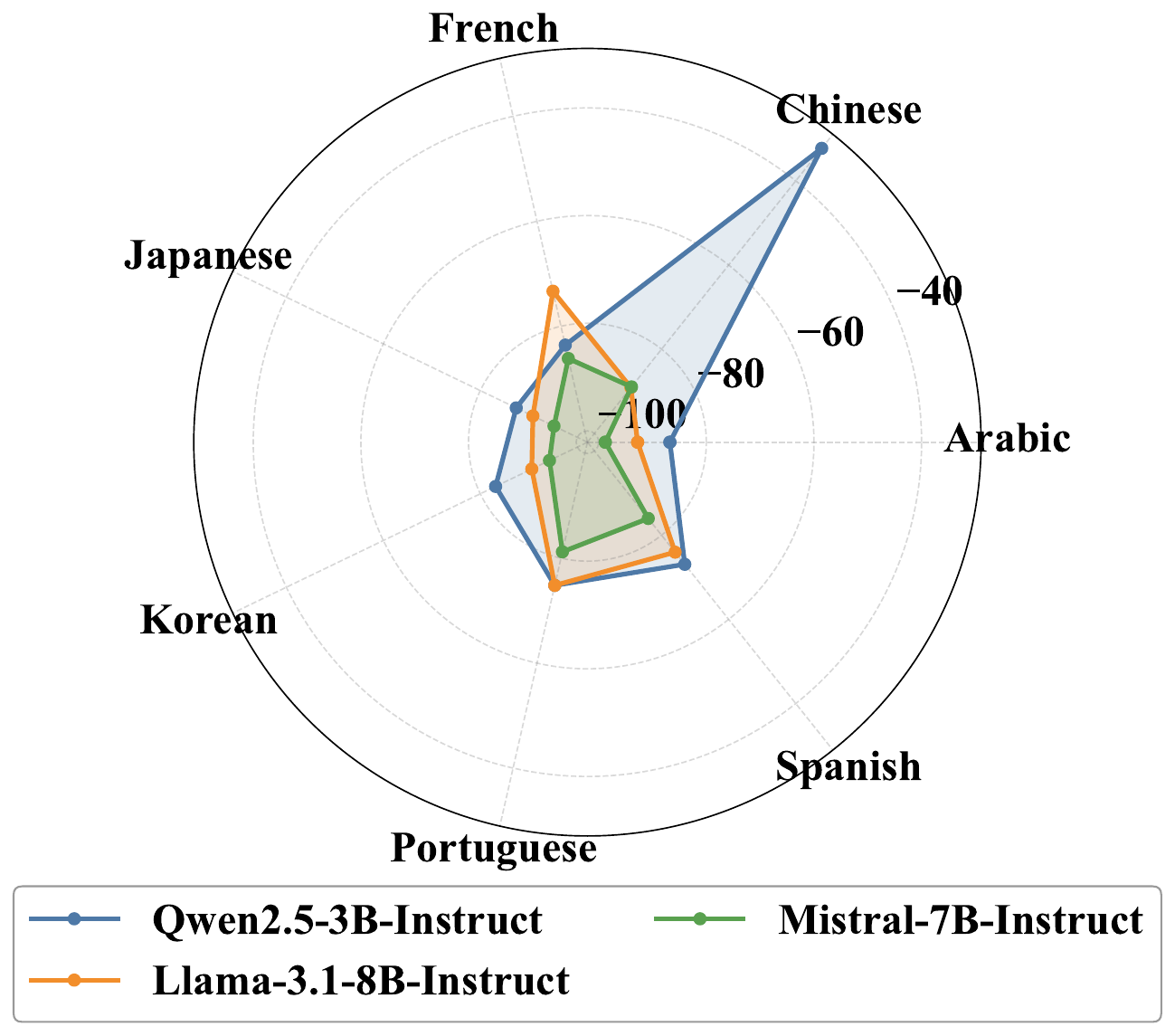}
    \end{subfigure}
    \begin{subfigure}{0.45\textwidth}
        \includegraphics[width=\textwidth]{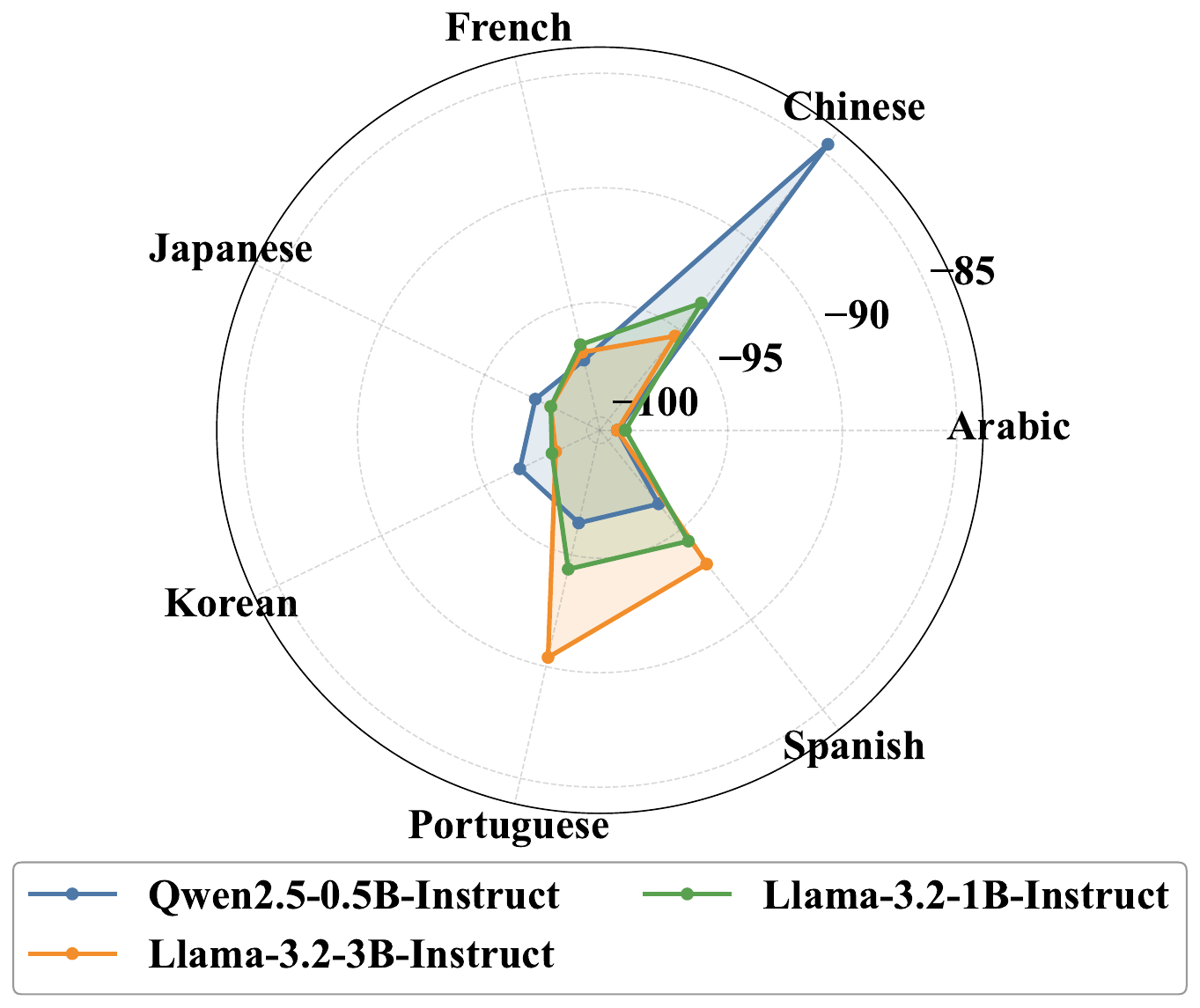}
    \end{subfigure}
    \vspace{-2mm}
    \caption{Net win rates (\%) of models under 3B parameters compared to the baseline model across languages.}
    \vspace{-3mm}
    \label{fig:rate_lang_all_2}
\end{figure*}

\begin{figure*}[!t]
    \centering
    \begin{subfigure}{0.45\textwidth}
        \includegraphics[width=\textwidth]{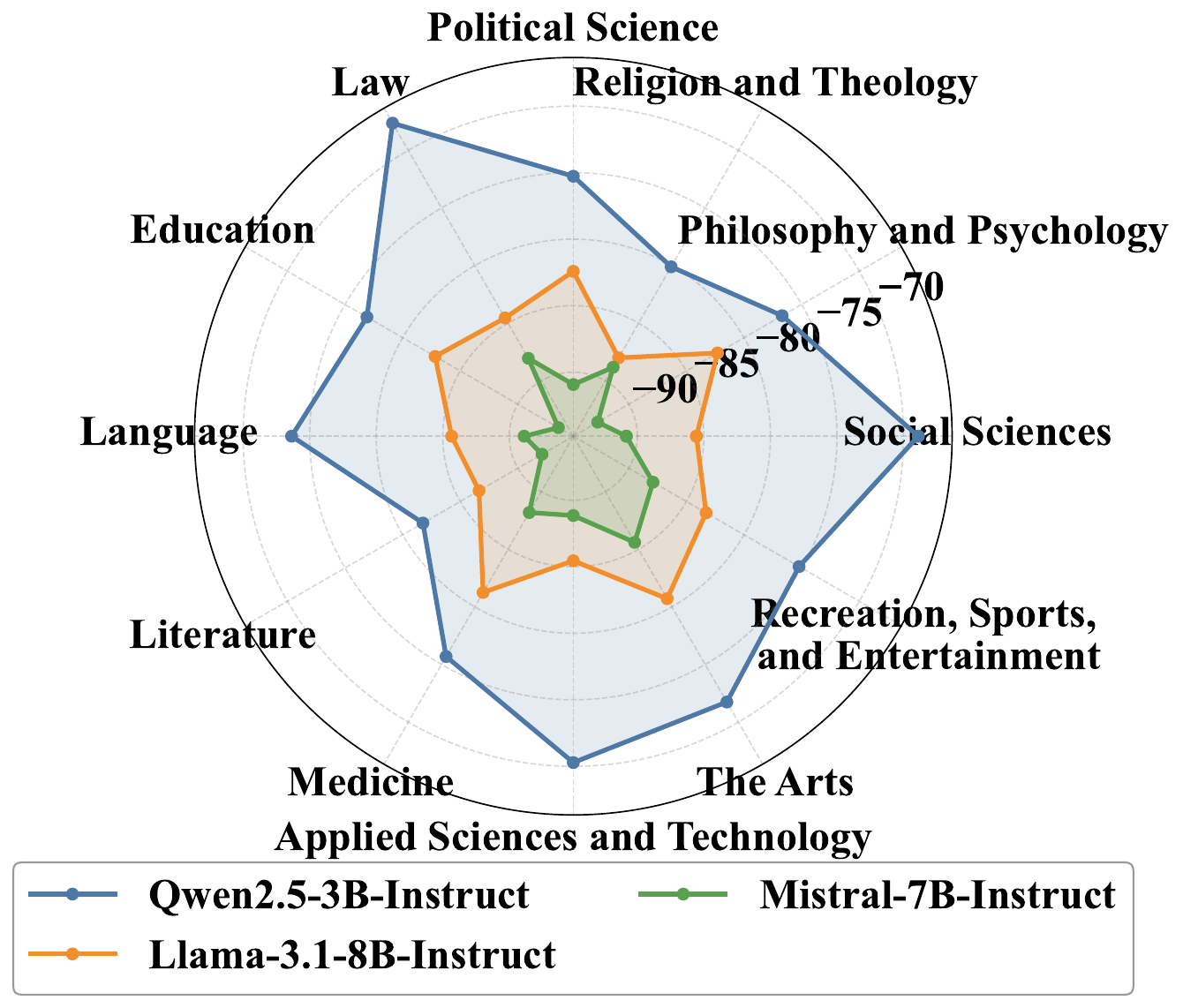}
        \caption{Third-tier cultural competence models}
    \end{subfigure}
    \begin{subfigure}{0.45\textwidth}
        \includegraphics[width=\textwidth]{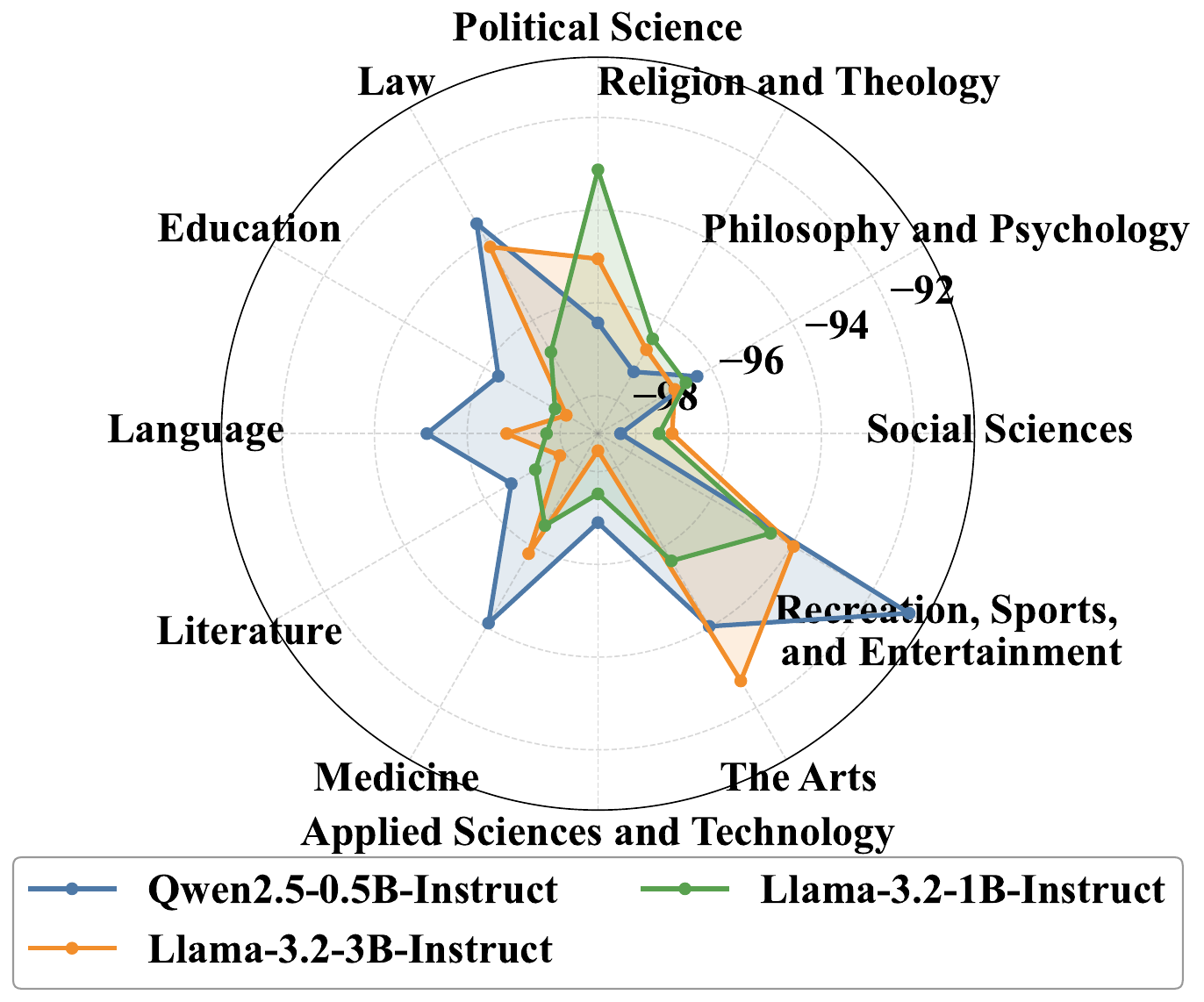}
        \caption{Fourth-tier cultural competence models}
    \end{subfigure}
    \vspace{-2mm}
    \caption{Net win rates (\%) of different models compared to the baseline model across cultural topics.}
    \vspace{-3mm}
    \label{fig:rate_topic_all_2}
\end{figure*}

Figures~\ref{fig:rate_lang_size} to Figure~\ref{fig:rate_topic_all_2} present net win rates (\%) across languages and cultural topics, comparing models of different sizes against the baseline.

\begin{table}[]
\centering
\small
\caption{Agreement rates between ChatGPT-4o-Latest and Qwen3‑32B‑think judges across all models}
\vspace{-2mm}
\begin{tabular}{lc}
\toprule
Model                      & Strict Agreement \\
\midrule
ChatGPT-4o-Latest          & 81.13\%          \\
GPT-4o mini                & 70.21\%          \\
Claude-3.5-Sonnet          & 75.08\%          \\
Qwen2.5-72B-Instruct       & 70.35\%          \\
Qwen2.5-7B-Instruct        & 78.38\%          \\
Qwen2.5-3B-Instruct        & 86.89\%          \\
Qwen2.5-0.5B-Instruct      & 97.06\%          \\
Llama-3.1-70B-Instruct     & 85.15\%          \\
Llama-3.1-8B-Instruct      & 90.96\%          \\
Llama-3.2-3B-Instruct      & 97.53\%          \\
Llama-3.2-1B-Instruct      & 97.08\%          \\
Mixtral-8x7B-Instruct-v0.1 & 84.60\%          \\
Mistral-Nemo-Instruct-2407 & 82.86\%          \\
Mistral-7B-Instruct-v0.3   & 93.42\%          \\
\midrule
Average                    & 85.05\%          \\
\bottomrule
\end{tabular}
\label{tab:agreement_rate}
\vspace{-3mm}
\end{table}

\begin{table*}[!t]
\footnotesize
\centering
\setlength\tabcolsep{4pt}
\caption{Net win rates (\%) of models across languages, relative to the Gemma2-27B-it baseline under the Qwen3‑32B‑Think judge. {\color{blue}Blue} and {\color{yellow}yellow} indicate the best and second-best performing models, respectively.}
\vspace{-2mm}
\begin{tabular}{lcccccccccccc}
\toprule  
{\color[HTML]{1F2329} Model}                      & Arabic       & Chinese       & French       & Japanese       & Korean       & Portuguese       & Spanish       & Weighted Average \\
\midrule
ChatGPT-4o-Latest &
  \cellcolor[HTML]{E1EAFF}68.61 &
  \cellcolor[HTML]{E1EAFF}79.93 &
  \cellcolor[HTML]{E1EAFF}80.85 &
  \cellcolor[HTML]{E1EAFF}65.22 &
  \cellcolor[HTML]{E1EAFF}71.18 &
  \cellcolor[HTML]{E1EAFF}80.26 &
  \cellcolor[HTML]{E1EAFF}88.94 &
  \cellcolor[HTML]{E1EAFF}76.33 \\
GPT-4o mini                & -34.04 & -26.42 & -9.57  & -29.54 & -27.26 & -18.61   & -1.28  & -20.92         \\
\midrule
Claude-3.5-Sonnet          & -51.68 & -21.91 & 7.52   & -9.75  & -31.77 & -14.10   & 19.07  & -14.08         \\
\midrule
Qwen2.5-72B-Instruct &
  \cellcolor[HTML]{FAF1D1}-0.35 &
  \cellcolor[HTML]{FAF1D1}-3.18 &
  \cellcolor[HTML]{FAF1D1}28.38 &
  \cellcolor[HTML]{FAF1D1}6.30 &
  \cellcolor[HTML]{FAF1D1}-4.69 &
  \cellcolor[HTML]{FAF1D1}16.17 &
  \cellcolor[HTML]{FAF1D1}32.37 &
  \cellcolor[HTML]{FAF1D1}10.80 \\
Qwen2.5-7B-Instruct        & -75.13 & -31.44 & -50.43 & -65.97 & -72.92 & -51.13   & -45.51 & -56.04         \\
Qwen2.5-3B-Instruct        & -94.71 & -60.37 & -87.18 & -85.91 & -83.51 & -85.53   & -82.85 & -82.77         \\
Qwen2.5-0.5B-Instruct      & -99.12 & -93.31 & -98.63 & -98.20 & -97.57 & -98.31   & -97.76 & -97.54         \\
\midrule
Llama-3.1-70B-Instruct     & -89.59 & -78.76 & -64.44 & -73.01 & -77.60 & -73.87   & -57.37 & -73.29         \\
Llama-3.1-8B-Instruct      & -95.41 & -89.30 & -85.13 & -88.89 & -91.32 & -85.69   & -83.65 & -88.45         \\
Llama-3.2-3B-Instruct      & -99.65 & -96.32 & -97.54 & -99.10 & -98.44 & -93.96   & -98.02 & -97.64         \\
Llama-3.2-1B-Instruct      & -99.29 & -95.82 & -97.44 & -98.65 & -98.26 & -96.24   & -94.87 & -97.23         \\
\midrule
Mixtral-8x7B-Instruct-v0.1 & -92.06 & -87.46 & -57.95 & -93.40 & -82.99 & -70.49   & -54.17 & -77.08         \\
Mistral-Nemo-Instruct-2407 & -93.47 & -55.02 & -81.37 & -67.77 & -67.19 & -80.08   & -81.89 & -74.98         \\
Mistral-7B-Instruct-v0.3   & -99.12 & -92.64 & -88.55 & -97.15 & -95.49 & -85.53   & -87.02 & -92.31        
\\
\bottomrule
\end{tabular}
\label{tab:net_win_rate_lang_1}
\vspace{-3mm}
\end{table*}

\begin{table*}[!t]
\footnotesize
\setlength\tabcolsep{2pt}
\caption{Net win rates (\%) of models across cultural topics, relative to the Gemma2-27B-it baseline under the Qwen3‑32B‑Think judge. {\color{blue}Blue} and {\color{yellow}yellow} indicate the best and second-best performing models, respectively.}
\vspace{-2mm}
\resizebox{\linewidth}{!}{
\begin{tabular}{lcccccccccccc}
\toprule  
Model & \begin{tabular}[c]{@{}c@{}}Social \\ Sciences\end{tabular} & \begin{tabular}[c]{@{}c@{}}Philosophy \\ and \\ Psychology\end{tabular} & \begin{tabular}[c]{@{}c@{}}Religion \\ and \\ Theology\end{tabular} & \begin{tabular}[c]{@{}c@{}}Political \\ Science\end{tabular} & Law & \begin{tabular}[c]{@{}c@{}}Edu-\\cation\end{tabular} & \begin{tabular}[c]{@{}c@{}}Lang-\\uage\end{tabular} & \begin{tabular}[c]{@{}c@{}}Liter-\\ature\end{tabular} & \begin{tabular}[c]{@{}c@{}}Medi-\\cine\end{tabular} & \begin{tabular}[c]{@{}c@{}}Applied \\ Sciences and \\ Technology\end{tabular} & \begin{tabular}[c]{@{}c@{}}Arts\end{tabular}   & \begin{tabular}[c]{@{}c@{}}Recreation, \\ Sports, and \\ Entertainment\end{tabular} \\
\midrule
ChatGPT-4o-Latest &
  \cellcolor[HTML]{E1EAFF}75.00 &
  \cellcolor[HTML]{E1EAFF}82.87 &
  \cellcolor[HTML]{E1EAFF}77.93 &
  \cellcolor[HTML]{E1EAFF}73.08 &
  \cellcolor[HTML]{E1EAFF}74.05 &
  \cellcolor[HTML]{E1EAFF}71.27 &
  \cellcolor[HTML]{E1EAFF}79.66 &
  \cellcolor[HTML]{E1EAFF}82.98 &
  \cellcolor[HTML]{E1EAFF}67.01 &
  \cellcolor[HTML]{E1EAFF}77.40 &
  \cellcolor[HTML]{E1EAFF}80.98 &
  \cellcolor[HTML]{E1EAFF}72.33 \\
GPT-4o mini                & -25.00 & -9.55    & -18.53 & -22.53  & -14.87 & -27.04  & -21.78 & -23.40   & -22.92 & -26.93 & -24.18 & -14.99   \\
\midrule
Claude-3.5-Sonnet          & -19.72 & -14.04   & -26.98 & -18.41  & -12.24 & -23.38  & -8.31  & -10.03   & -5.56  & -14.24 & -2.17  & -11.53   \\
\midrule
Qwen2.5-72B-Instruct &
  \cellcolor[HTML]{FAF1D1}14.17 &
  \cellcolor[HTML]{FAF1D1}18.82 &
  \cellcolor[HTML]{FAF1D1}12.53 &
  \cellcolor[HTML]{FAF1D1}7.14 &
  \cellcolor[HTML]{FAF1D1}7.29 &
  \cellcolor[HTML]{FAF1D1}4.23 &
  \cellcolor[HTML]{FAF1D1}14.04 &
  \cellcolor[HTML]{FAF1D1}15.20 &
  \cellcolor[HTML]{FAF1D1}15.28 &
  \cellcolor[HTML]{FAF1D1}1.24 &
  \cellcolor[HTML]{FAF1D1}11.96 &
  \cellcolor[HTML]{FAF1D1}7.78 \\
Qwen2.5-7B-Instruct        & -52.50 & -62.92   & -55.59 & -55.77  & -53.64 & -59.72  & -57.59 & -62.92   & -43.06 & -53.56 & -59.24 & -53.60   \\
Qwen2.5-3B-Instruct        & -86.11 & -81.46   & -83.38 & -85.71  & -82.51 & -84.51  & -84.81 & -85.11   & -82.64 & -81.42 & -79.08 & -76.37   \\
Qwen2.5-0.5B-Instruct      & -98.33 & -97.19   & -97.28 & -98.08  & -96.79 & -98.59  & -97.99 & -98.48   & -97.57 & -97.83 & -97.01 & -95.39   \\
\midrule
Llama-3.1-70B-Instruct     & -80.00 & -73.60   & -76.29 & -70.88  & -66.18 & -78.59  & -75.07 & -76.90   & -70.49 & -71.83 & -70.92 & -68.01   \\
Llama-3.1-8B-Instruct      & -93.06 & -91.57   & -89.10 & -86.54  & -88.63 & -91.83  & -87.97 & -92.10   & -86.46 & -87.00 & -81.79 & -85.22   \\
Llama-3.2-3B-Instruct      & -98.30 & -98.30   & -98.62 & -96.67  & -96.63 & -98.58  & -97.98 & -97.23   & -97.20 & -99.38 & -97.27 & -95.63   \\
Llama-3.2-1B-Instruct      & -98.33 & -97.19   & -96.19 & -97.80  & -96.21 & -99.44  & -97.13 & -98.18   & -98.96 & -95.67 & -95.65 & -96.25   \\
\midrule
Mixtral-8x7B-Instruct-v0.1 & -87.50 & -78.93   & -71.39 & -74.45  & -76.97 & -80.56  & -76.50 & -77.20   & -75.35 & -78.02 & -75.27 & -72.62   \\
Mistral-Nemo-Instruct-2407 & -79.72 & -75.56   & -75.48 & -73.08  & -69.97 & -74.93  & -74.79 & -75.08   & -72.22 & -72.45 & -81.79 & -73.49   \\
Mistral-7B-Instruct-v0.3   & -95.00 & -91.29   & -94.01 & -92.31  & -90.96 & -94.93  & -90.83 & -95.14   & -91.32 & -91.64 & -91.30 & -88.76  \\
\midrule
\cellcolor[HTML]{DEE0E3}Average &
  \cellcolor[HTML]{DEE0E3}-58.89 &
  \cellcolor[HTML]{DEE0E3}-54.99 &
  \cellcolor[HTML]{DEE0E3}-56.60 &
  \cellcolor[HTML]{DEE0E3}-56.57 &
  \cellcolor[HTML]{DEE0E3}-54.59 &
  \cellcolor[HTML]{DEE0E3}-59.76 &
  \cellcolor[HTML]{DEE0E3}-55.50 &
  \cellcolor[HTML]{DEE0E3}-56.69 &
  \cellcolor[HTML]{DEE0E3}-54.39 &
  \cellcolor[HTML]{DEE0E3}-56.52 &
  \cellcolor[HTML]{DEE0E3}-54.48 &
  \cellcolor[HTML]{DEE0E3}-53.70
\\
\bottomrule
\end{tabular}
}
\label{tab:net_win_rate_topic_1}
\vspace{-3mm}
\end{table*}

\section{Prompt Templates}
\label{sec:prompt}
\vspace{-1mm}

Figure~\ref{fig:prompt1} shows the prompt template for cultural topic extension. Figures~\ref{fig:prompt2} to \ref{fig:prompt8} are the step-by-step prompt templates for QA pairs generation. And Figure~\ref{fig:prompt9} is the prompt template for pairwise comparison of different model responses.


\begin{figure*}[t]
\begin{tcolorbox}[width=\textwidth]\footnotesize
Assuming you are an {\color{red}expert in the field of \{primary\_topics\} in \{country\}}, below are possible cultural differences in primary topics and secondary topics. For each secondary topic, evaluate whether there are differences between \{country\} and other countries worldwide. If there are no differences, output "None"; if there are differences, expand the secondary topic into five or more tertiary topics and keywords that {\color{red}reflect the unique characteristics of \{country\}} and cover as much content as possible.\\

\# Primary Topics - Secondary Topics:\\
\{primary\_topics\} - \{secondary\_topics\} \\

{\color{red}Please don't be lazy and answer this question in depth from a local's perspective. }
\end{tcolorbox}
\vspace{-3mm}
\caption{Prompt for extending primary and secondary topics into tertiary topics and keywords.}
\label{fig:prompt1}
\end{figure*}

\begin{figure*}[t]
\begin{tcolorbox}[width=\textwidth]\footnotesize
Can you extract knowledge points related to the "\{keyword\}" from the following text? \\
Yes or No. {\color{red}Do not output other content.}\\

\# Text (Title: \{wikipedia\_title\})\\
\{wikipedia\_content\}

\end{tcolorbox}
\vspace{-3mm}
\caption{Prompt for determining whether the retrieved page is related to the keyword (Step 1).}
\label{fig:prompt2}
\end{figure*}

\begin{figure*}[t]
\begin{tcolorbox}[width=\textwidth]\footnotesize
Assume you are an {\color{red}expert in the \{primary\_topic\}}. Please determine whether the following text contains multicultural differences or is specific cultural knowledge unique to \{country\}. {\color{red}Choose A, B, or C as your output. Do not output other content.} \\

A. Contains cultural differences from different countries \\
B. Is cultural knowledge specific to \{country\} \\
C. Neither of the above \\

\# Text (Title: \{title\})\\
\{content\}

\end{tcolorbox}
\vspace{-3mm}
\caption{Prompt for determining whether the retrieved page contains culture-specific content (Step 1).}
\label{fig:prompt3}
\end{figure*}

\begin{figure*}[t]
\begin{tcolorbox}[width=\textwidth]\footnotesize
From the following references, extract up to 3 points of knowledge in \{target\_language\} that are important, diverse, and reflect the differences between \{country\} and other countries.

\# References (Title: \{title\}) \\
\{content\} \\

{\color{red}\# Requirements} \\
- First, summarize knowledge points for {country} by starting with "In \{country\},". \\
- Then, summarize the knowledge points for countries other than {country} and different from {country} accordingly, starting with "In [other countries],". If there are no knowledge points for countries other than {country}, explain how the USA is different based on your knowledge, starting with "In [US],". \\
- Do not extract historically relevant knowledge points. \\
- Return in JSON format: {[}\{"knowledge1": "xxx", "differ1": "xxx"\}, \{"knowledge2": "xxx", "differ2": "xxx"\}, \{"knowledge3": "xxx", "differ3": "xxx"\}{]} \\
- Use \{target\_language\}. \\
\end{tcolorbox}
\vspace{-3mm}
\caption{Prompt for knowledge extraction in different cultural knowledge settings (Step 2).}
\label{fig:prompt4}
\end{figure*}

\begin{figure*}[t]
\begin{tcolorbox}[width=\textwidth]\footnotesize
Based on the following references, extract no more than 3 text fragments from the references, which contain cultural knowledge unique to \{country\}. \\

\# References (Title: {title}) \\
\{content\}  \\

{\color{red}\# Requirements} \\
- Do not extract historical or non-\{country\}-related knowledge points. \\
- Return in JSON format: {[}\{"knowledge1": "xxx"\}, \{"knowledge2": "xxx"\}, \{"knowledge3": "xxx"\}{]} \\
- Use \{target\_language\}.
\end{tcolorbox}
\vspace{-3mm}
\caption{Prompt for knowledge extraction in unique cultural knowledge settings (Step 2).}
\label{fig:prompt5}
\end{figure*}

\begin{figure*}[t]
\begin{tcolorbox}[width=\textwidth]\footnotesize
Assuming you are an {\color{red}expert in the field of \{primary\_topic\}}, pose a situational question in \{target\_language\} that involves the following different knowledge points from two countries. \\

{\color{red}\# Example} \\
\#\# Knowledge \\
In Japan, 8 (\jp{八}): pronounced similarly to the word for prosperity or development (\jp{はち}, hachi). 7 (\jp{七}): considered a lucky number, symbolizing good things come in pairs, and celebrated in many traditional festivals such as Shichi-Go-San (Children's Day). 4 (\jp{四}): pronounced similar to the word for death (\jp{し}, shi). 9 (\jp{九}): pronounced similar to the word for suffering (\jp{く}, ku), meaning pain and hardship. 
\\
\#\# Different Knowledge \\
In the United States, 7 (seven): widely regarded as a lucky number, often appears in gambling, lottery, and other occasions, such as 7 on Las Vegas slot machines. 3 (three): In Western culture, there is a concept of "three", such as "threesome", "lucky three", and is considered a perfectly balanced number. 13 (thirteen): considered unlucky, in many buildings, the 13th floor is even skipped and directly numbered as the 14th floor. This superstition comes from "The Last Supper", in which Judas is the thirteenth person. 666: considered to be the "devil's number", from the Book of Revelation in the Bible. In China, 8 (\zh{八}), pronounced similar to "fa" (the "fa" in "facai"), symbolizes wealth and success, and is deeply loved by people. 6 (\zh{六}): also represents good luck. 9 (\zh{九}): symbolizes longevity and longevity. 4 (\zh{四}): pronounced similar to "si" (\zh{死}), is considered an unlucky number.
\\
\{\{"Question": "Can you recommend a few lucky numbers?"\}\} \\

{\color{red}\# Your Turn} \\
\#\# Knowledge \\
\{knowledge\} \\
\#\# Different Knowledge \\
\{differ\} \\

{\color{red}\# Requirements} \\
- The question must not contain any offensive or discriminatory content, nor should it include pornography, gruesomeness, violence, or aggressive elements. \\
- Don't mention country names in the question. \\
- Do not use referential words, demonstrative pronouns, or demonstrative pronouns. \\
- Directly output the question, do not provide other contents. \\
- Use \{target\_language\}. 
\end{tcolorbox}
\vspace{-3mm}
\caption{Prompt for question generation in different cultural knowledge settings (Step 3).}
\label{fig:prompt6}
\end{figure*}

\begin{figure*}[t]
\begin{tcolorbox}[width=\textwidth]\footnotesize
Consider the following cultural knowledge and assume specific scenarios or roles to {\color{red}seek expert advice} in \{target\_language\}.

\# Cultural Knowledge \\
\{knowledge\} \\

{\color{red}\# Requirements} \\
- The question must not contain any offensive or discriminatory content, nor should it include pornography, gruesomeness, violence, or aggressive elements. \\
- The question can be asked in different cultural contexts. \\
- Don't mention country names in the question. \\
- Do not use referential words or demonstrative pronouns. \\
- Directly output the question, do not provide other contents. \\
- Use \{target\_language\}.

\end{tcolorbox}
\vspace{-3mm}
\caption{Prompt for question generation in unique cultural knowledge settings (Step 3).}
\label{fig:prompt7}
\end{figure*}

\begin{figure*}[t]
\begin{tcolorbox}[width=\textwidth]\footnotesize
You are an {\color{red}expert in the field of \{primary\_topic\} in \{country\}}. Please refer to the following cultural knowledge to answer the questions.

\# Cultural Knowledge \\
\{knowledge\}  \\

\# Question \\
\{questinon\} \\

{\color{red}\# Requirements} \\
- Be as detailed as possible and closely follow the cultural knowledge provided. \\
- Be clear, detailed, and address multiple aspects of the question comprehensively. \\
- Don't mention country names in the answer. \\
- Do not use referential words or demonstrative pronouns. \\
- Directly output the answer; do not provide other contents. \\
- Use \{target\_language\}. \\
\end{tcolorbox}
\vspace{-3mm}
\caption{Prompt for answer generation in both different and unique cultural knowledge settings (Step 3).}
\label{fig:prompt8}
\end{figure*}

\begin{figure*}[t]
\begin{tcolorbox}[width=\textwidth]\footnotesize
{[}System{]}\\
Please act as an impartial judge and evaluate the quality of the responses provided by two AI assistants to the user question displayed below. Your evaluation should consider correctness and helpfulness. You will be given a reference answer, assistant A's answer, and assistant B's answer. Your job is to evaluate which assistant's answer is better. Begin your evaluation by comparing both assistants' answers with the reference answer. Identify and correct any mistakes. Avoid any position biases and ensure that the order in which the responses were presented does not influence your decision. Do not allow the length of the responses to influence your evaluation. Do not favor certain names of the assistants. Be as objective as possible. After providing your explanation, output your final verdict by strictly following this format: "{[}{[}A{]}{]}" if assistant A is better, "{[}{[}B{]}{]}" if assistant B is better, and "{[}{[}C{]}{]}" for a tie.\\
\\
{[}User Question{]}\\
\{question\}\\
\\
{[}The Start of Reference Answer{]}\\
\{answer\_ref\}\\
{[}The End of Reference Answer{]}\\
\\
{[}The Start of Assistant A's Answer{]}\\
\{answer\_a\}\\
{[}The End of Assistant A's Answer{]}\\
\\
{[}The Start of Assistant B's Answer{]}\\
\{answer\_b\}\\
{[}The End of Assistant B's Answer{]}
\end{tcolorbox}
\vspace{-3mm}
\caption{Prompt for reference-guided pairwise comparison~\cite{zheng2023judging}.}
\label{fig:prompt9}
\end{figure*}

\section{Universal Cultural Topics}

Table~\ref{tab:topics_all} presents the complete list of primary and secondary topics.
 
\begin{table*}[!t]
\footnotesize
\caption{Universal Cultural Topics}
\vspace{-2mm}
\begin{tabular}{|l|l|l|}
\hline
\textbf{Abbreviations} & \textbf{Primary Topics} & \textbf{Secondary Topics} \\ \hline
\multirow{21}{*}{SS} & Social Sciences & Methods of the social sciences \\ \cline{2-3} 
 & Social Sciences & Social questions. Social practice \\ \cline{2-3} 
 & Social Sciences & Cultural practice \\ \cline{2-3} 
 & Social Sciences & Way of life (Lebensweise) \\ \cline{2-3} 
 & Social Sciences & Food, Beverage, and Culinary Arts \\ \cline{2-3} 
 & Social Sciences & Gender studies \\ \cline{2-3} 
 & Social Sciences & \begin{tabular}[c]{@{}l@{}}Sociography. Descriptive studies of society \\ (both qualitative and quantitative)\end{tabular} \\ \cline{2-3} 
 & Social Sciences & Statistics as a science. Statistical theory \\ \cline{2-3} 
 & Social Sciences & Society \\ \cline{2-3} 
 & Social Sciences & Economics. Economic science \\ \cline{2-3} 
 & Social Sciences & Public administration. Government. Military affairs \\ \cline{2-3} 
 & Social Sciences & Safeguarding the mental and material necessities of life \\ \cline{2-3} 
 & Social Sciences & Costume. Clothing. National dress. Fashion. Adornment \\ \cline{2-3} 
 & Social Sciences & Customs, manners, usage in private life \\ \cline{2-3} 
 & Social Sciences & Death. Treatment of corpses. Funerals. Death rites \\ \cline{2-3} 
 & Social Sciences & Public life. Pageantry. Social life. Life of the people \\ \cline{2-3} 
 & Social Sciences & Social ceremonial. Etiquette. Good manners. Social forms. Rank. Title \\ \cline{2-3} 
 & Social Sciences & Folklore in the strict sense \\ \cline{2-3} 
 & Social Sciences & Commerce \\ \cline{2-3} 
 & Social Sciences & Demography \\ \cline{2-3} 
 & Social Sciences & Social Interaction \\ \hline
\multirow{10}{*}{PP} & Philosophy and Psychology & Metaphysics \\ \cline{2-3} 
 & Philosophy and Psychology & Epistemology, causation and humankind \\ \cline{2-3} 
 & Philosophy and Psychology & Parapsychology and occultism \\ \cline{2-3} 
 & Philosophy and Psychology & Specific philosophical schools and viewpoints \\ \cline{2-3} 
 & Philosophy and Psychology & Psychology \\ \cline{2-3} 
 & Philosophy and Psychology & Philosophical logic \\ \cline{2-3} 
 & Philosophy and Psychology & Ethics (Moral philosophy) \\ \cline{2-3} 
 & Philosophy and Psychology & Ancient, medieval and eastern philosophy \\ \cline{2-3} 
 & Philosophy and Psychology & Modern western and other noneastern philosophy \\ \cline{2-3} 
 & Philosophy and Psychology & Other philosophy and psychology \\ \hline
\multirow{10}{*}{RT} & Religion and Theology & Prehistoric religions. Religions of early societies \\ \cline{2-3} 
 & Religion and Theology & Religions originating in the Far East \\ \cline{2-3} 
 & Religion and Theology & \begin{tabular}[c]{@{}l@{}}Religions originating in Indian sub-continent. \\ Hindu religion in the broad sense\end{tabular} \\ \cline{2-3} 
 & Religion and Theology & Buddhism \\ \cline{2-3} 
 & Religion and Theology & Religions of antiquity. Minor cults and religions \\ \cline{2-3} 
 & Religion and Theology & Judaism \\ \cline{2-3} 
 & Religion and Theology & Christianity \\ \cline{2-3} 
 & Religion and Theology & Islam \\ \cline{2-3} 
 & Religion and Theology & Modern spiritual movements \\ \cline{2-3} 
 & Religion and Theology & Other religions \\ \hline
\multirow{12}{*}{PS} & Political Science & Political science (General) \\ \cline{2-3} 
 & Political Science & Political theory. Theory of the state \\ \cline{2-3} 
 & Political Science & Political institutions and public administration \\ \cline{2-3} 
 & Political Science & North America \\ \cline{2-3} 
 & Political Science & United States \\ \cline{2-3} 
 & Political Science & Canada, Latin America, etc. \\ \cline{2-3} 
 & Political Science & Europe \\ \cline{2-3} 
 & Political Science & Asia \\ \cline{2-3} 
 & Political Science & Local government. Municipal government \\ \cline{2-3} 
 & Political Science & \begin{tabular}[c]{@{}l@{}}Colonies and colonization. Emigration and immigration. \\ International migration\end{tabular} \\ \cline{2-3} 
 & Political Science & International relations \\ \cline{2-3} 
 & Political Science & Other Politics and Policy \\ \hline
\multirow{10}{*}{LAW} & Law & Law in general. Comparative and uniform law. Jurisprudence \\ \cline{2-3} 
 & Law & Religious law \\ \cline{2-3} 
 & Law & United Kingdom and Ireland law \\ \cline{2-3} 
 & Law & Canada law \\ \cline{2-3} 
 & Law & United States law \\ \cline{2-3} 
 & Law & Europe law \\ \cline{2-3} 
 & Law & Germany law \\ \cline{2-3} 
 & Law & Asia and Eurasia law \\ \cline{2-3} 
 & Law & Africa law \\ \cline{2-3} 
 & Law & Latin America law \\ \hline
\end{tabular}
\label{tab:topics_all}
\end{table*}

\begin{table*}[]
\footnotesize
\begin{tabular}{|l|l|l|}
\hline
\textbf{Abbreviations} & \textbf{Primary Topics} & \textbf{Secondary Topics} \\ \hline
\multirow{2}{*}{LAW} & Law & Law of nations \\ \cline{2-3} 
 & Law & Other laws \\ \hline
\multirow{5}{*}{EDU} & Education & Education (General) \\ \cline{2-3} 
 & Education & History of education \\ \cline{2-3} 
 & Education & Theory and practice of education \\ \cline{2-3} 
 & Education & Special aspects of education \\ \cline{2-3} 
 & Education & Other educational aspects \\ \hline
\multirow{16}{*}{LAN} & Language & Linguistics \\ \cline{2-3} 
 & Language & English and Old English (Anglo-Saxon) \\ \cline{2-3} 
 & Language & German and related languages \\ \cline{2-3} 
 & Language & French and related Romance languages \\ \cline{2-3} 
 & Language & Italian, Dalmatian, Romanian, Rhaetian, Sardinian, Corsican \\ \cline{2-3} 
 & Language & Spanish, Portuguese, Galician \\ \cline{2-3} 
 & Language & Latin and related Italic languages \\ \cline{2-3} 
 & Language & Classical Greek and related Hellenic languages \\ \cline{2-3} 
 & Language & Chinese, Cantonese \\ \cline{2-3} 
 & Language & Arabic \\ \cline{2-3} 
 & Language & Russian \\ \cline{2-3} 
 & Language & Japanese \\ \cline{2-3} 
 & Language & Vietnamese \\ \cline{2-3} 
 & Language & Thai \\ \cline{2-3} 
 & Language & Korean \\ \cline{2-3} 
 & Language & Other languages \\ \hline
\multirow{16}{*}{LIT} & Literature & American literature in English \\ \cline{2-3} 
 & Literature & English and Old English (Anglo-Saxon) literatures \\ \cline{2-3} 
 & Literature & German literature and literatures of related languages \\ \cline{2-3} 
 & Literature & French literature and literatures of related Romance languages \\ \cline{2-3} 
 & Literature & \begin{tabular}[c]{@{}l@{}}Literatures of Italian, Dalmatian, Romanian, Rhaetian, \\ Sardinian, Corsican languages\end{tabular} \\ \cline{2-3} 
 & Literature & Literatures of Spanish, Portuguese, Galician languages \\ \cline{2-3} 
 & Literature & Latin literature and literatures of related Italic languages \\ \cline{2-3} 
 & Literature & \begin{tabular}[c]{@{}l@{}}Classical Greek literature and literatures of related \\ Hellenic languages\end{tabular} \\ \cline{2-3} 
 & Literature & Literatures of Chinese, Cantonese \\ \cline{2-3} 
 & Literature & Literatures of Arabic \\ \cline{2-3} 
 & Literature & Literatures of Russian \\ \cline{2-3} 
 & Literature & Literatures of Japanese \\ \cline{2-3} 
 & Literature & Literatures of Vietnamese \\ \cline{2-3} 
 & Literature & Literatures of Thai \\ \cline{2-3} 
 & Literature & Literatures of Korean \\ \cline{2-3} 
 & Literature & Literatures of other specific languages and language families \\ \hline
MED & MEDICINE & Medical sciences \\ \hline
\multirow{13}{*}{AST} & Applied Sciences and Technology & Biotechnology \\ \cline{2-3} 
 & Applied Sciences and Technology & Engineering. Technology in general \\ \cline{2-3} 
 & Applied Sciences and Technology & \begin{tabular}[c]{@{}l@{}}Agriculture and related sciences and techniques. Forestry. \\ Farming. Wildlife exploitation\end{tabular} \\ \cline{2-3} 
 & Applied Sciences and Technology & Home economics. Domestic science. Housekeeping \\ \cline{2-3} 
 & Applied Sciences and Technology & Transportation and communications \\ \cline{2-3} 
 & Applied Sciences and Technology & Accountancy \\ \cline{2-3} 
 & Applied Sciences and Technology & Business management \\ \cline{2-3} 
 & Applied Sciences and Technology & Public relations \\ \cline{2-3} 
 & Applied Sciences and Technology & Chemical technology. Chemical and related industries \\ \cline{2-3} 
 & Applied Sciences and Technology & Various industries, trades and crafts \\ \cline{2-3} 
 & Applied Sciences and Technology & Industries, crafts and trades for finished or assembled articles \\ \cline{2-3} 
 & Applied Sciences and Technology & \begin{tabular}[c]{@{}l@{}}Building (construction) trade. Building materials. \\ Building practice and procedure\end{tabular} \\ \cline{2-3} 
 & Applied Sciences and Technology & Intelligent Technology \\ \hline
\multirow{11}{*}{ART} & Arts & Special auxiliary subdivision for the arts \\ \cline{2-3} 
 & Arts & \begin{tabular}[c]{@{}l@{}}Physical planning. Regional, town and country planning. \\ Landscapes, parks, gardens\end{tabular} \\ \cline{2-3} 
 & Arts & Architecture \\ \cline{2-3} 
 & Arts & Plastic arts \\ \cline{2-3} 
 & Arts & Drawing. Design. Applied arts and crafts \\ \cline{2-3} 
 & Arts & Painting \\ \cline{2-3} 
 & Arts & Graphic art, printmaking. Graphics \\ \cline{2-3} 
 & Arts & Photography and similar processes \\ \cline{2-3} 
 & Arts & Music and dance \\ \cline{2-3} 
 & Arts & Drama and Movie \\ \cline{2-3} 
 & Arts & Other arts \\ \hline
\end{tabular}
\end{table*}

\begin{table*}[!t]
\footnotesize
\begin{tabular}{|l|l|l|}
\hline
\textbf{Abbreviations} & \textbf{Primary Topics} & \textbf{Secondary Topics} \\ \hline
\multirow{3}{*}{RSE} & Recreation, Sports, and Entertainment & Games \\ \cline{2-3} 
 & Recreation, Sports, and Entertainment & Sport and Exercise \\ \cline{2-3} 
 & Recreation, Sports, and Entertainment & Other entertainment, leisure\hspace{4cm} \\ \hline
\end{tabular}
\end{table*}

\end{document}